\documentclass[preprint,authoryear,12pt]{elsarticle}

\usepackage{graphicx}
\usepackage{subfigure}
\usepackage{float}
\usepackage{natbib}

\usepackage{rotating}

\usepackage{epsfig}
\usepackage{epstopdf}
\usepackage{amssymb}
 \usepackage{amsthm}
\usepackage{array}
\usepackage{lineno}
\usepackage{longtable}
\usepackage{amsmath}
\usepackage{float}
\usepackage{bm}
\usepackage{rotating}
\usepackage{tabularx}
\usepackage{mathrsfs}
\usepackage{hyperref}

\usepackage{caption}
\journal{ISPRS J. Photogramm. Remote Sens.}

\begin{document}

\begin{frontmatter}
\title{Efficiently utilizing complex-valued PolSAR image data via a multi-task deep learning framework}

\author{Lamei Zhang}
\ead{lmzhang@hit.edu.cn}

\author{Hongwei Dong}
\ead{donghongwei1994@163.com}

\author{Bin Zou\corref{cor1}}
\cortext[cor1]{Corresponding author}
\address{Department of Information Engineering, Harbin Institute of Technology, Harbin 150001, China}
\ead{zoubin@hit.edu.cn}

\begin{abstract}
Convolutional neural networks (CNNs) have been widely used to improve the accuracy of polarimetric synthetic aperture radar (PolSAR) image classification. However, in most studies, the difference between PolSAR images and optical images is rarely considered. Most of the existing CNNs are not tailored for the task of PolSAR image classification, in which complex-valued PolSAR data have been simply equated to real-valued data to fit the optical image processing architectures and avoid complex-valued operations. This is one of the reasons CNNs unable to perform their full capabilities in PolSAR classification. To solve the above problem, the objective of this paper is to develop a tailored CNN framework for PolSAR image classification, which can be implemented from two aspects: Seeking a better form of PolSAR data as the input of CNNs and building matched CNN architectures based on the proposed input form. In this paper, considering the properties of complex-valued numbers, amplitude and phase of complex-valued PolSAR data are extracted as the input for the first time to maintain the integrity of original information while avoiding immature complex-valued operations. Then, a multi-task CNN (MCNN) architecture is proposed to match the improved input form and achieve better classification results. Furthermore, depthwise separable convolution is introduced to the proposed architecture in order to better extract information from the phase information. Experiments on three PolSAR benchmark datasets not only prove that using amplitude and phase as the input do contribute to the improvement of PolSAR classification, but also verify the adaptability between the improved input form and the well-designed architectures.
\end{abstract}

\begin{keyword}
Deep learning \sep Convolutional neural networks \sep Polarimetric synthetic aperture radar (PolSAR) classification \sep Two-stream architecture \sep Multi-task learning \sep Depthwise separable convolutions
\end{keyword}

\end{frontmatter}

\section{Introduction}
\label{sec:1}
Polarimetric synthetic aperture radar (PolSAR), as one of the most advanced detectors in the field of remote sensing, can comprehensively describe the targets in all-weather and all-times. PolSAR can capture rich information from surface of the Earth so that it has a wide range of applications in various fields, such as agriculture, fishery, urban planning, environmental monitoring, etc. Classification can be seen as the basic step of PolSAR image interpretation, which has been widely concerned by researchers for a long time. In recent years, the accuracy of PolSAR image classification has made great progress with the maturity of pattern recognition methods. However, due to the complexity of echo imaging mechanism, efficiently mining knowledge from complex-valued PolSAR data is still an open problem.
\par In the past few years, deep learning \citep{Deep} based methods have achieved significant progress on a variety of tasks. Different from the manually designed \citep{Zou2017Independent} or statistical learning based feature engineering methods \citep{PCA}, deep learning has a automated feature engineering process accomplished by a deep neural network. It can find more abstract representations than classical hand-craft methods from the original data. There are two main reasons for the popularity of deep learning. The first is high degree of flexibility, which reflects in the ability to fit any forms of the input data, such as image \citep{Lecun2014Backpropagation}, natural language \citep{word2vec}, audio \citep{Graves2013Speech} and video \citep{Shuiwang20133D}. For different kinds of input, good results can be achieved by changing network architectures according to the characteristics of the data to be processed, which avoids the difficulty of designing hand-craft feature extractors. The second is powerful capacity of feature extraction. Deep neural networks can extract high-level features which cannot be obtained by traditional methods and more generalized features naturally avoid the difficulty of designing multi-class classifiers. Obviously, such an end-to-end learning framework is suitable for improving PolSAR image classification.
\par Deep learning based image processing models, represented by convolutional neural networks (CNNs), are the focus of our attention since ImageNet Large-Scale Visual Recognition Challenge 2012 \citep{Krizhevsky2012ImageNet}. At present, CNN-based methods can deal with various multi-class classification problems \citep{ResNet,DenseNet}. Pioneering works which used CNNs to interpret SAR or PolSAR images have existed for some time. \citet{Chen2016Target} used a chain-structured CNN to classify SAR images for the first time, which embodied the strong ability of CNNs for feature extraction and achieved good results on MSTAR dataset. Deep features and shallow features were combined, and a fully convolutional classifier was proposed in \citet{Yan2018A} for better performance. As another form of deep learning, autoencoders have also been used for PolSAR image classification \citep{Hou2017Classification,AE1,AE2}. Some scholars sought higher accuracy by developing more powerful network architectures or better combination of hyperparameters \citep{CNN1,CNN2,CNN3}. In addition, CNNs also demonstrated potential capabilities in SAR image denoising \citep{Despeckling1}, remote sensing image registration \citep{Despeckling2} and pan-sharpening \citep{Despeckling3}.
\par Although CNNs-based PolSAR image classification methods have been developed to some extent, few studies designed their network architectures according to the characteristics of PolSAR images. Unlike optical images, the echo imaging mechanism of PolSAR brings richer information as well as complex-valued data format. The scattering properties of PolSAR images can be fully described by the complex-valued polarization scattering matrix, of which each element represents the backscattering coefficients produced by the received polarized electromagnetic waves from different directions. According to the properties of complex-valued numbers, the amplitude and phase information can be transformed and obtained based on complex-valued PolSAR data. The amplitude information corresponds to the backscattering intensity of the electromagnetic wave from the target to the radar, which has a great correlation with the gray scale information obtained by visible light imaging. The phase information corresponds to the distance between the sensor platform and the target, which is not available from other detectors. Moreover, there is a coupling relationship between the phase information obtained from different transmitting and receiving directions of electromagnetic wave. That is why PolSAR can better reflect the scattering properties of targets than SAR. However, abundant information and distinctive nature also make PolSAR image interpretations difficult. Manually designed feature extractors combined with efficient classifiers have been the mainstream approaches of PolSAR classification in the previous methods. With the great success of deep learning in computer vision area, how to improve PolSAR classification with the aid of deep learning techniques is an urgent problem to be solved.
\par From the author's point of view, the main problem needs to be solved when applying deep learning to PolSAR image classification is: Differences between PolSAR images and optical images should be profoundly considered. It is well known that CNNs originate from computer vision and optical image processing, and most CNNs only consider extracting features from amplitude information. Many studies in PolSAR area followed the pipeline of extracting information from a single angle, which ignore the unique phase information of PolSAR images. We believe that the existing CNN architectures should be used in a targeted way, rather than blindly following.
\par In order to develop a efficient deep learning based framework for PolSAR image classification, we consider two key issues in this work: Seeking better input forms and building tailored network architectures. Firstly, we concern about what form of PolSAR data should be input into the CNNs. It is intuitive to construct a complex-valued neural network to fit the complex-valued PolSAR data. However, the research of complex-valued neural networks are currently marginal and most of complex-valued neural networks still stay in conjectures. Therefore, most studies in PolSAR area treated the complex-valued source data as real-valued and directly split it into real and imaginary parts as the input of CNNs. This is forced to make sacrifices in order to take advantage of the optical image processing models due to the lack of PolSAR tailored architectures. Thus, one of the objectives of this paper is to seek a representation form which not only preserves the information of complex-valued PolSAR data maximally, but also avoids the complex-valued operations. Another key issue is the design of CNN architectures. According to the connotation of deep representation learning, network architectures should be adaptively adjusted for the different inputs. However, it has been neglected in most applications. A matched network architecture is necessary for better classification when improved input forms have been studied. In addition, the potential coupling relationship of the PolSAR phase information should be considered to better identify the targets. Above all, the other objective of this paper is to design matched CNN architectures to fit the improved input form for better classification results.
\par Some related studies have made efforts to design exclusive architectures for PolSAR image classification, which have great inspirations for our work. \citet{complex} proposed a three-layers complex-valued CNN in order to fit the PolSAR data as well as achieve better classification. A 3D convolutions based CNN was proposed to better extract the relationships between the channels of PolSAR images \citep{3dcnn}. Besides, \citet{input1} used hand-craft PolSAR features as the input of CNNs to improve the performance. A polarimetric scattering coding method was proposed in \citet{input2} to get a new form of input, and a fully convolutional network based classifier was used to obtain better classification results.
\par Inspired by the previous works, a deep learning based PolSAR image classification framework is proposed in this paper based on the proposed input form and multi-task CNN (MCNN). Amplitude and phase of complex-valued PolSAR data are extracted as the input of CNNs, which can fully retain the information of complex-valued data under the premise of avoiding complex-valued operations. To the best of our knowledge, this is the first work which explores the significance of input forms for CNNs-based PolSAR classification methods. Moreover, a multi-task CNN is developed to better match the input data. During the process of network modeling, Two-stream \citep{twostream1}, deeply supervised \citep{Lee2014Deeply} and densely connected architectures \citep{DenseNet} are integrated, considering two parts information (amplitude and phase) of the input. Further, depthwise separable convolution \citep{xception} is introduced to better extract information from the phase of PolSAR images. Experimental results on three PolSAR benchmark datasets demonstrate the validity of the proposed classification methods.
\par The rest of this paper is organized as follows: Background and relevant technical literature are reviewed in \autoref{sec:2}. The proposed methods and relevant analyses are listed in \autoref{sec:3}. In \autoref{sec:4}, experimental results are exhibited. The comparison between the proposed classification framework and some other ones is shown in \autoref{sec:add}. Conclusion and possible future directions are given in \autoref{sec:5}.
\section{Background}
\label{sec:2}
In this section, we briefly review the background of involved researches.
\subsection{CNNs}
\label{sec:2.1}
CNNs have been widely used in the field of image processing and achieved the state-of-the-arts in a variety of tasks, such as image classification \citep{inception,DenseNet}, semantic segmentation \citep{Shelhamer2014Fully,Ronneberger2015U,Badrinarayanan2017SegNet}, instance segmentation \citep{He2017Mask}, target detection \citep{Ren2015Faster,Redmon2015You,Liu2015SSD}, fine-grained recognition \citep{Fu2017Look}, etc. It greatly changed the traditional pipeline of image processing methods and established an end-to-end feature engineering and recognition framework. A continuous non-convex optimization problem is solved according to the network architecture and the objective function to seek a mapping from the original data to the manual labels. Unlike hand-craft feature extractors, the success of CNNs is largely attributed to their automation of the feature engineering process. Although more automated sub-fields are emerging \citep{nas1,nas2}, the design of network architectures and the construction of objectives incorporate the wisdom of the human experts. CNNs have the ability to utilize massive data compared to the shallow models in machine learning \citep{Cortes1995Support,WANG201941,Zhang2015Fully}. Moreover, the generation of fast computing technology based on graphics processing unit (GPU) greatly promotes the application of CNNs. For a long time, CNNs have followed the LeNet-style \citep{Lenet5} architecture with cascaded layers, which start from a set of primitives including convolutions, pooling, fully connected and classifier. Convolution layer is used to extract features. Pooling layer is used to increase the receptive field as well as reduce the computational complexity. Fully connected layer is used to change the feature dimension to cooperate with softmax classifier to achieve multi-class classification.
\par Many works have been extensively studied to improve the performance of CNNs. Rectified linear unit (ReLU) nonlinear activation function \citep{Nair2010Rectified}, batch normalization (BN) \citep{Ioffe2015Batch} and skip connection \citep{ResNet} are commonly used tricks to increase the depth of CNNs for better generalization performance. Relevant methods for initializing network parameters can be seen in \citep{par1,par2}. Some works pursue wider instead of deeper architectures, and have achieved good results \citep{inception}. Some variants of the vanilla convolutions have been studied to adapt to different tasks, such as 3D convolutions \citep{3dcnn,c3d}, dilated convolutions \citep{yudilated}, depthwise separable convolutions \citep{xception,mnet}, and group convolutions \citep{Krizhevsky2012ImageNet,ResNext}. In addition to maximum pooling and average pooling, some new methods have been developed \citep{Bruna2014Signal,Zeiler2013Stochastic,Kaiming2014Spatial,linnin}. Alternative activation functions like Leaky ReLU \citep{LRelu}, PReLu \citep{par2}, ELU \citep{Xu2015Empirical} have also been studied for better performance.

\subsection{CNNs for PolSAR classification}
\label{sec:2.2}
\par A lot of PolSAR image classification methods have been developed based on CNNs \citep{CNN1,CNN2,CNN3,CNN4,CNN5}. The aim of PolSAR image classification is to give a certain category to every pixel, which can be seen as a semantic segmentation problem in computer vision.

\begin{figure}[h]
\begin{centering}
\includegraphics[width=12.0cm,height=4.3cm]{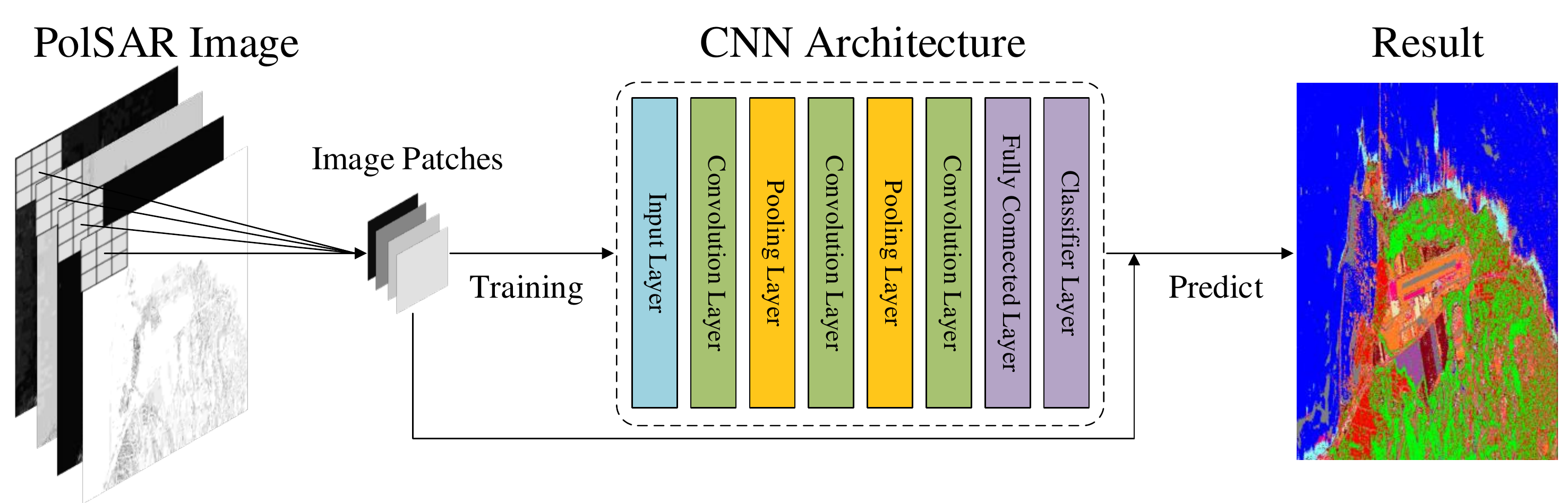}
\caption{CNNs for PolSAR image classification. Different types of layers are visualized by different colors. All elements of the complex-valued PolSAR scattering matrix (or covariance, coherency matrix) are divided into their real parts and imaginary parts, and each decomposed value is equally treated as a channel of the input. Image patches are obtained according to manual labels as inputs to train the network. Then, each pixel enters the trained model to get the classification result of whole map.}\label{fig2.2}
\end{centering}
\end{figure}

Fully convolutional networks (FCNs) \citep{Shelhamer2014Fully} have been proposed to solve semantic segmentation problems, which classify each pixel of the image simultaneously. FCNs can avoid the loss of detail information and shorten the running time. However, having massive manually labels is a prerequisite for the implementation of FCNs, which is less likely to be accomplished in the context of PolSAR image processing. Therefore, PolSAR image classification still follows the pattern of slicing image first and then recognizing all the patches, which can be seen from \autoref{fig2.2}.
\par Studies on CNNs-based PolSAR image classification are in the ascendant \citep{CNN2,CNN3,CNN6}. However, most of them did not consider how to design a tailored architecture to fit PolSAR data, but directly followed the backbones of optical image classification. Considering the differences between PolSAR images and optical images, it is advisable to adjust the network architectures to better utilize the unique information of PolSAR. Hand-craft PolSAR features were extracted as the input of CNNs in some studies \citep{input1,input2}, which are attempts in the right direction. Real-valued primitives operations in CNNs were replaced by complex-valued ones in \citet{complex} and a three-layers complex-valued CNN was modeled. However, the development of complex-valued neural networks is not mature, and many tricks of network designing have not been extended to the complex domain.
\subsection{Two-stream CNNs}
\label{sec:2.3}
\par Multi-stream CNNs were studied in order to cope with video processing \citep{twostream1,twostream2}. The core idea of this kind of models is using multiple branches to process each types of multi-source data respectively. They are based on the hypothesis that the data in different types may be difficult to extract typical features together. It have been found that these architectures which increase the width of networks can also improve the performance of image processing \citep{Guo2016Two,inception}. Although optical images only have amplitude information, there are many other kinds of images whose data can be divided into more than one type, especially in the field of remote sensing \citep{6844831,7018910}. Moreover, such architectures are of great significance for multi-source information fusion \citep{Hu2017FusioNet} and multimodal deep learning \citep{7789544}.
\begin{figure}[!h]
\begin{centering}
\includegraphics[width=\textwidth,height=6.0cm]{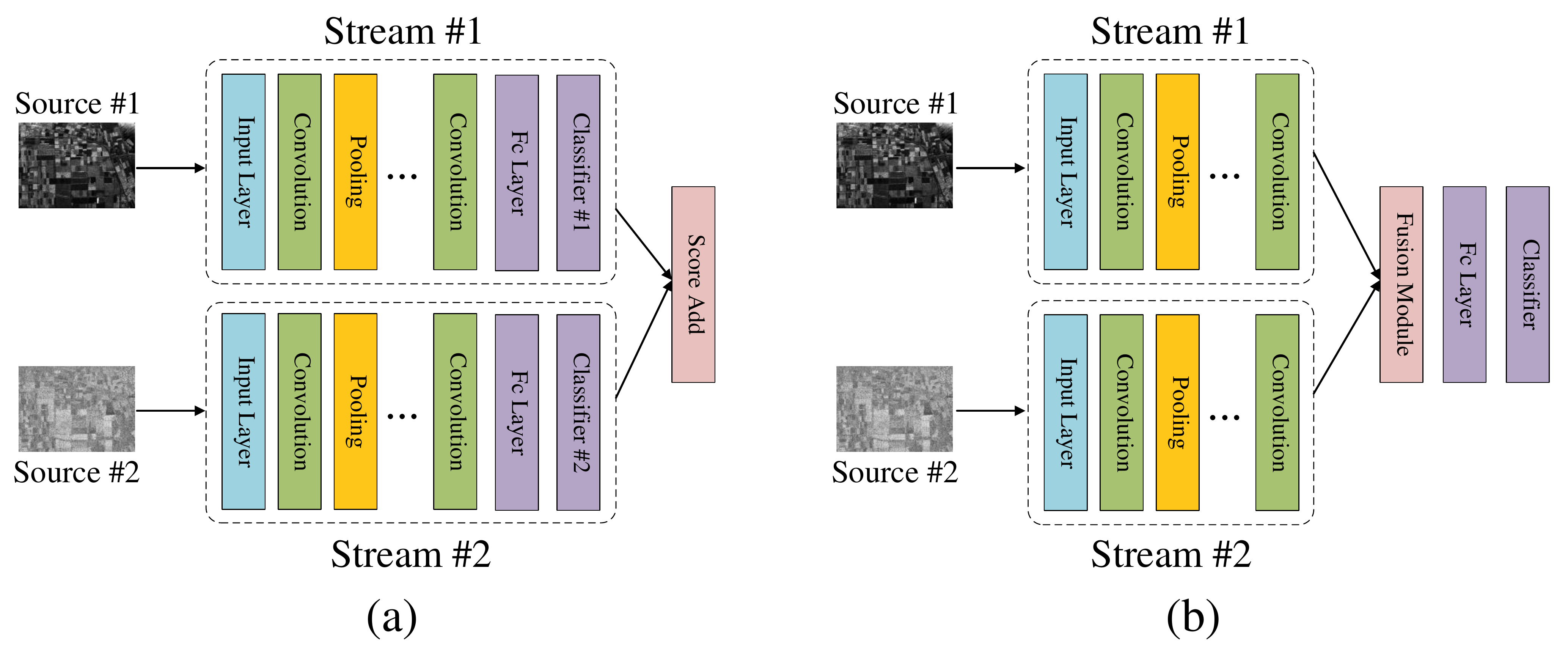}
\caption{Architectures of two-stream CNN. (a) Two-stream CNN without information interaction. (b) Improved two-stream CNN with information fusion module.}
\label{fig1.1}
\end{centering}
\end{figure}
\par As a special case of multi-stream architectures, ordinary two-stream CNN \citep{twostream1} and its improved version with a information fusion module \citep{Guo2016Two,twostream2} can be seen in \autoref{fig1.1}. The two architectures in \autoref{fig1.1} are similar, and the difference lies in whether the two branches have information interaction or not. If there is no interaction between the branches, the usual practice is to average the classification possibilities of the two branches. Sum, max, concatenation, convolution and bilinear pooling can be used to seek higher-order feature fusion \citep{twostream2}.
\subsection{Deeply-supervised architectures}
\label{sec:2.4}
Deeply-supervised architectures were studied in \citep{Lee2014Deeply} to discover the capabilities of the middle layers. Initially, its goal was to improve the performance and prevent gradients vanishing \citep{inception}. Later, a series of studies show that it can also enhance semantic segmentation and edge detection models \citep{Xie2015Holistically,Hou2016Deeply}.
\begin{figure}[!h]
\begin{centering}
\includegraphics[width=12.0cm,height=4.6cm]{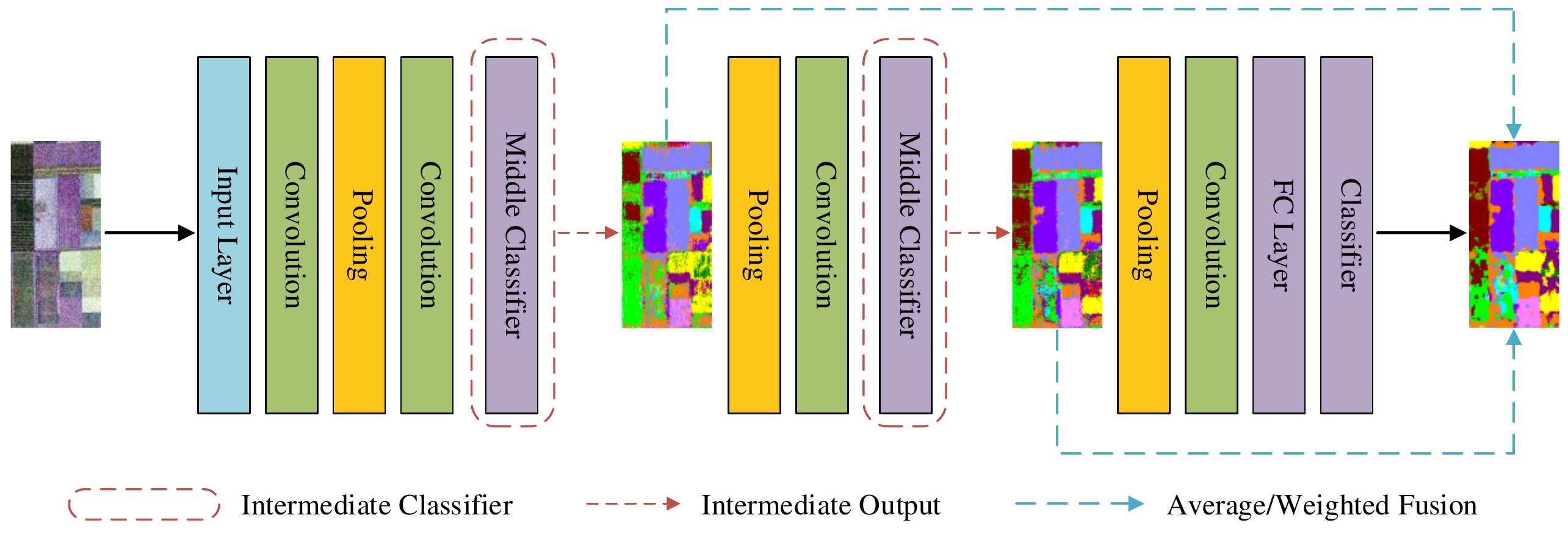}
\caption{Deeply-supervised architectures for semantic segmentation or edge detection. Intermediate outputs of the middle layers are utilized for better performance.}
\label{fig2.4}
\end{centering}
\end{figure}
\par As shown in \autoref{fig2.4}, the validity of deeply-supervised architectures is due to the utilization of middle-level output. Many existing studies have proved that the features extracted from deep parts of a architecture are global, while shallow and middle parts are local \citep{Hou2016Deeply}. This inspires us to pay more attention to the intermediate output in the task of PolSAR image classification.
\section{Methods}
\label{sec:3}
In this section, we present implementation details of the proposed classification framework. First, complex-valued PolSAR data is converted into the form of amplitude and phase. Then, some of labeled samples are used to train the proposed multi-task CNN (MCNN). After the process of training, all labeled samples are used to test the accuracy of the classification method. Finally, the whole image pixel-wise classification result is obtained based on the trained network.
\subsection{Representation of PolSAR data}
\label{sec:3.1}
Polarization scattering matrix can fully characterize the electromagnetic scattering properties of different types of ground targets. The scattering matrix is defined as:
\begin{equation}
\left[\begin{array}{c}
    S
\end{array}\right]=
\left[\begin{array}{cc}
    S_{HH} & S_{HV}      \\
    S_{VH} & S_{VV}
\end{array}\right],\label{1}
\end{equation}
where $S_{PQ}(P, Q\in \{H, V\})$ represents the backscattering coefficient of the polarized electromagnetic wave in emitting $Q$ direction and receiving $P$ direction. $H$ and $V$ represent the horizontal and vertical polarization respectively. According to the reciprocity theorem, the $S$ matrix satisfies $S_{HV}=S_{VH}$. In order to describe the scattering properties of targets more clearly, the $S$ matrix is usually vectorized, and the polarization coherence matrix or polarization covariance matrix are obtained. Polarization vector and polarization coherence matrix based on Pauli decomposition are expressed as Eqs. \eqref{2} and \eqref{3}
\begin{equation}
\vec{k}=\frac{1}{\sqrt{2}}[S_{HH}+S_{VV}, S_{HH}-S_{VV}, 2S_{HV}]^T,\label{2}
\end{equation}
\begin{equation}
[T]=\langle \vec{k}\vec{k}^{*T}\rangle.\label{3}
\end{equation}
\par The polarization coherence matrix $T$ is a Hermitian matrix, of which each element except the diagonal is complex-valued. Elements of the upper triangular matrix $[T_{11}, T_{12}, T_{13}, T_{22}, T_{23}, T_{33}]$ are commonly used as the input to CNNs. Thus, three real-valued numbers and three complex-valued numbers describe each pixel of a PolSAR image. The usual practice is to split the real and imaginary parts of the three complex-valued numbers to utilize the mainstream CNN backbones. At this point, there are nine numbers to describe each pixel if $T$ matrix is used as the input. Such preprocessing avoids complex-valued operations. However, the encapsulation of complex-valued data is broken. Notice that for a complex number $z=a+bi,(z\in\mathbb{C}, a,b\in\mathbb{R})$, it can be expressed in the form of $z=rexp(\varphi i),(r,\varphi\in\mathbb{R})$, where $r$ and $\varphi$ represent the amplitude and phase respectively. They can be calculated by the following formulas:
\begin{equation}
r(a,b)=\sqrt{a^2+b^2}\label{4}
\end{equation}
and
\begin{equation}
\varphi(a,b)=\left\{
\begin{array}{cc}
   \arctan\frac{b}{a}, &    a> 0     \\
     \pm \frac{\pi}{2}, &  a=0,b\neq0 \\
          \arctan\frac{b}{a}\pm\pi, &  a<0,b\neq0 \\
               \pi, &  a<0,b=0.
\end{array}\right.\label{5}
\end{equation}
\par Complex-valued PolSAR data can be converted into its corresponding amplitude and phase according to Eqs. \eqref{4} and \eqref{5}.
After the above operations, the input data which contains two different types of information is available. Since the diagonal elements of the covariance/coherency originally include amplitude information, six channels of PolSAR amplitude information and three channels of PolSAR phase information are obtained as the proposed input form.
\subsection{The proposed MCNN}
\label{sec:3.2}
\begin{figure}[h]
\begin{centering}
\includegraphics[width=12.0cm,height=6.0cm]{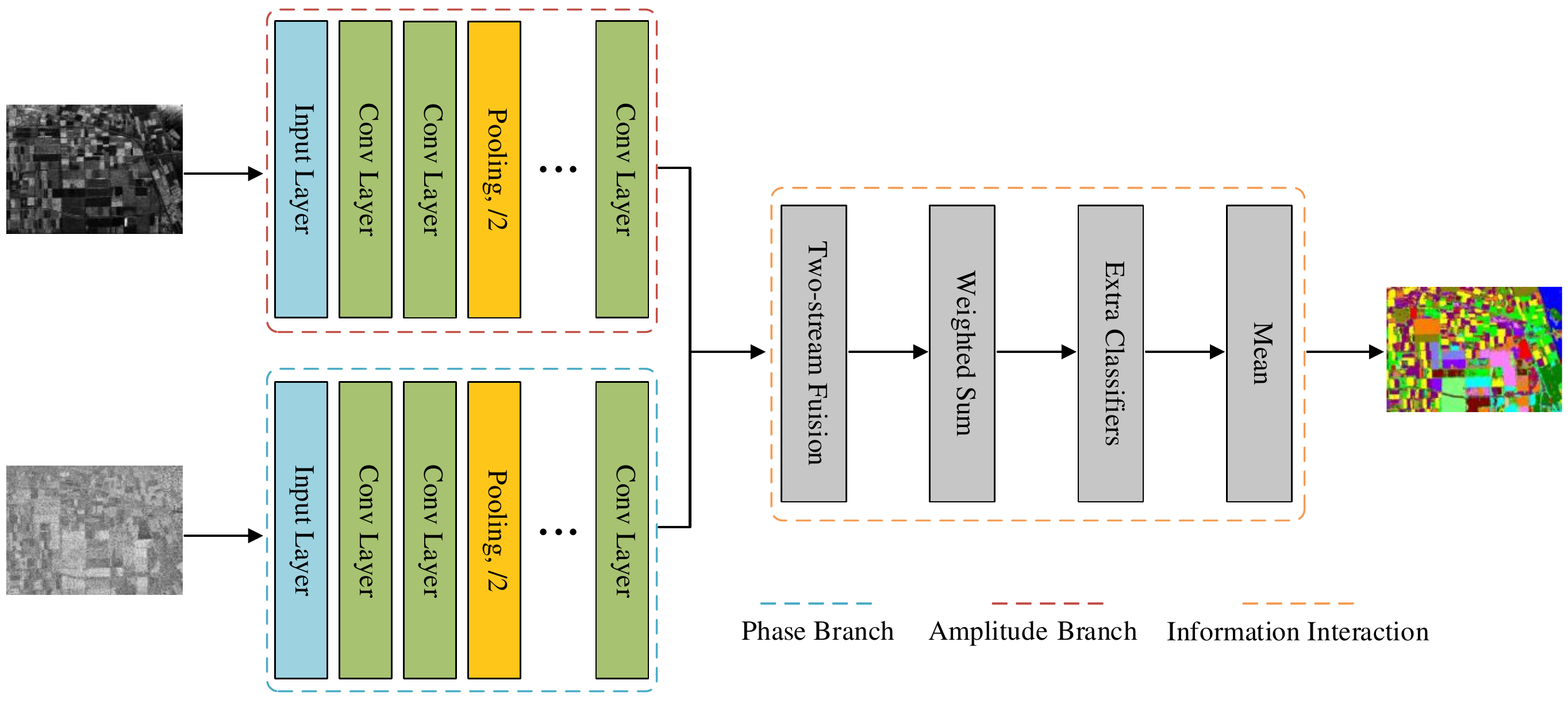}
\caption{General view of the proposed MCNN. The architecture consists of three parts: Phase branch (blue dotted line), amplitude branch (red dotted line) and interaction module (yellow dotted line). Two branches are used to extract features from different types of data respectively, which is adjusted specific to the input. The high-order interaction part is used to fuse to high-level features extracted from two branches. Such a well-designed fusion module is designed to make more effective use of the obtained higher-order features.}\label{fig4}
\end{centering}
\end{figure}

In order to maintain the integrity of complex-valued PolSAR data, complex-valued numbers are transformed into their amplitude and phase as the input of CNNs. Then, MCNN is proposed based on two-stream architectures \citep{Guo2016Two,twostream2} to fit the input form. The proposed architecture extracts features from both amplitude and phase respectively by different branches, and a carefully designed fusion module \citep{DenseNet} is used for information interaction. Moreover, side outputs are fully utilized to improve performance inspired by \citet{Xie2015Holistically,Hou2016Deeply}. The general view of the proposed architecture can be seen from \autoref{fig4}. Compared with the ordinary CNNs, more network branches are considered during the process of architecture modeling to match the improved input. Compared with the two-stream architectures in \autoref{fig1.1}, the proposed not only has better fusion mechanism, but also incorporates the idea of deeply-supervised networks to make better use of the side outputs. Next, we introduce the components of the architecture and their advantages.
\par
\begin{figure}[h]
\begin{centering}
\includegraphics[width=6.0cm,height=5.6cm]{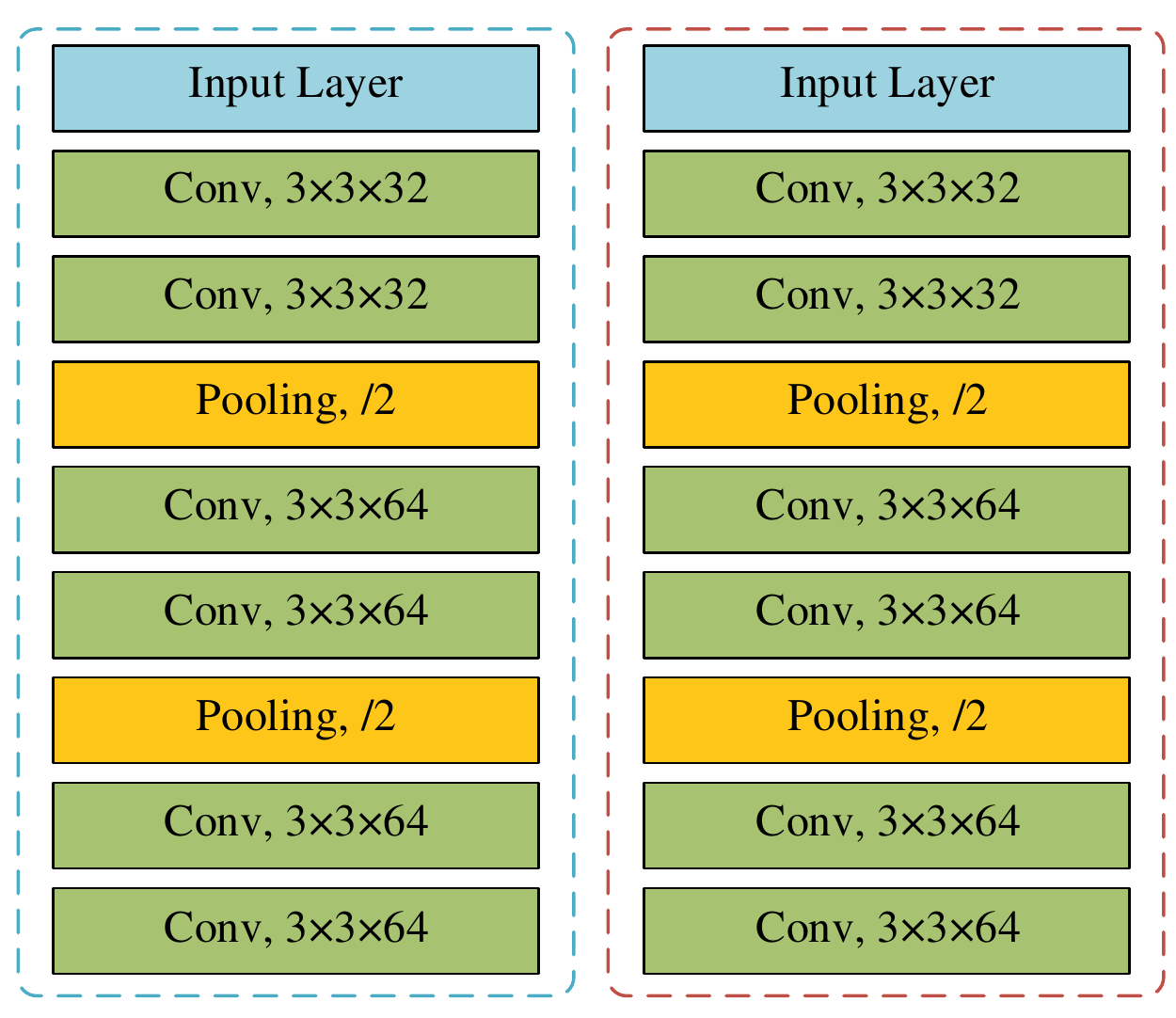}
\caption{Architecture of the branches of MCNN. Each branch consists of six trainable convolution layers and two pooling layers, and the hyperparameters are shared in both branches.}\label{fig_branch}
\end{centering}
\end{figure}

\textbf{Individual branch:} This part corresponds to the blue and red dotted boxes in \autoref{fig4}. Inspired by the two-stream architectures shown in \autoref{fig1.1}, two branches are designed as the main parts of MCNN to process the amplitude and phase of PolSAR images respectively. It is well known that the VGGNet \citep{vgg} improves the ordinary LeNet-style CNN architectures and achieves better performance. Therefore, a VGG-style backbone is adopted when designing the branches of MCNN. Details of the architecture can be seen from \autoref{fig_branch}.
\par Chain-structured architectures are used for individual branches, and each convolution layer in \autoref{fig2.2} is replaced by convolution blocks composed of two cascade convolution layers. To alleviate over-fitting, dropout mechanism \citep{Srivastava2014Dropout} is introduced which discards some neurons of its input during the process of training. Therefore, the output feature maps of convolution and fully connected operations will be forcibly set to zero at a ratio whose value will be determined later. After that, batch normalization layer and ReLU activation layer are employed (omitted in the figure for convenience) after each convolution operation. It is worth to note that such a VGGNet-based backbone with the composite style of \emph{Conv}$3\times3$-\emph{BN}-\emph{ReLU} has been shown to be effective in some advanced CNN classifiers \citep{ResNet,DenseNet}. Relatively high-level features can be extracted by individual branches with three convolution blocks and two pooling layers. For the setting of hyperparameters, 32 is set to the depth of the first convolution block and the one of the remaining two are set to be 64, which can be seen as a tradeoff between complexity and precision.
\par\textbf{Early fusion:} This part corresponds to the two-stream fusion module in \autoref{fig4}, We call it early fusion because another deeper fusion module exists. The two-stream architecture without information interaction is not recommended \citep{twostream2}. Although some operations including sum, max, concatenation, convolution and bilinear pooling, are considered to achieve the information exchange of the features obtained by different branches. Coarse fusion operations may result in the loss of information in higher-level features. A phenomenon can be observed that many works directly used the output of two branches and achieved qualified results \citep{twostream1,7018910}. Based on this, we can make a reasonable hypothesis: Output of the branches is beneficial to good recognition. Therefore, how to keep the obtained information while mining the more abstract one is what we need to consider in the process of building fusion modules.
\par Convolution undoubtedly has stronger non-linear fitting ability compared with basic operations. Therefore, it has become the mainstream choice for building fusion modules in many studies \citep{twostream2,Guo2016Two,Hu2017FusioNet}. However, vanilla convolution can not maintain the information of its input. Fortunately, feature reuse can be implemented through densely connected architectures \citep{DenseNet}, whose mathematical expression can be written as:
\begin{equation}
x_{l}=f_l([x_0,x_1,\cdots,x_{l-1}]),\label{6}
\end{equation}
where $x_i$ is output of the $i$th layer and $f_i(\cdot)$ is the operation of this layer.
\begin{figure}[h]
\begin{centering}
\includegraphics[width=12.0cm]{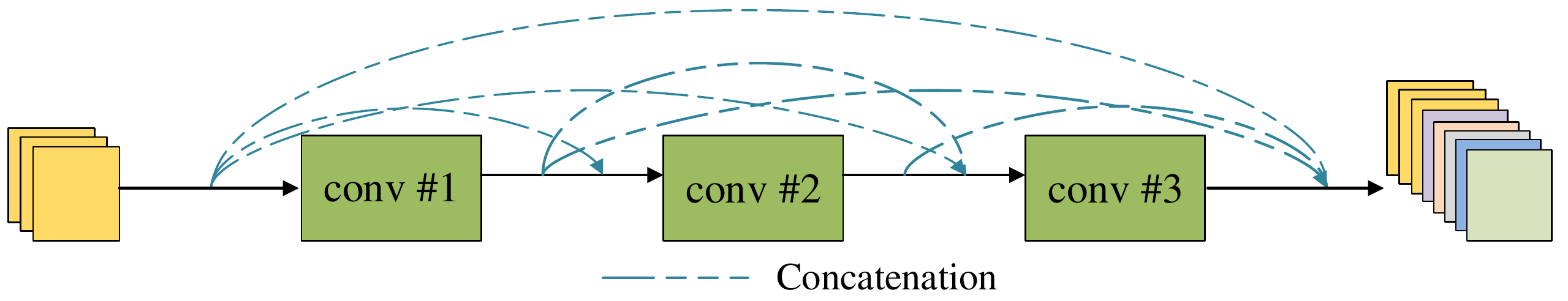}
\caption{A four layers Dense block, in which each layer takes all preceding feature maps as input.}\label{fig3}
\end{centering}
\end{figure}
\par As shown in \autoref{fig3}, convolution block with densely connected convolution layers can keep the feature maps obtained at each layer through skip connections and concatenate operations. It has been proved that such architectures can protect the existing high-level features and promote the feature reuse. Above all, a five layers dense block is used to replace the ordinary convolution fusion module to achieve better information interaction.
\par\textbf{Advanced fusion:} The fusion module of \autoref{fig1.1}b has been improved by early fusion layers, which can be seen as an interaction at the feature level. Inspired by the architectures shown in \autoref{fig2.4}, an advanced fusion mechanism at the level of classification results is sought by a weighted sum layer and three extra classifiers.
\par
\begin{figure}[h]
\begin{centering}
\includegraphics[width=8.0cm,height=5.0cm]{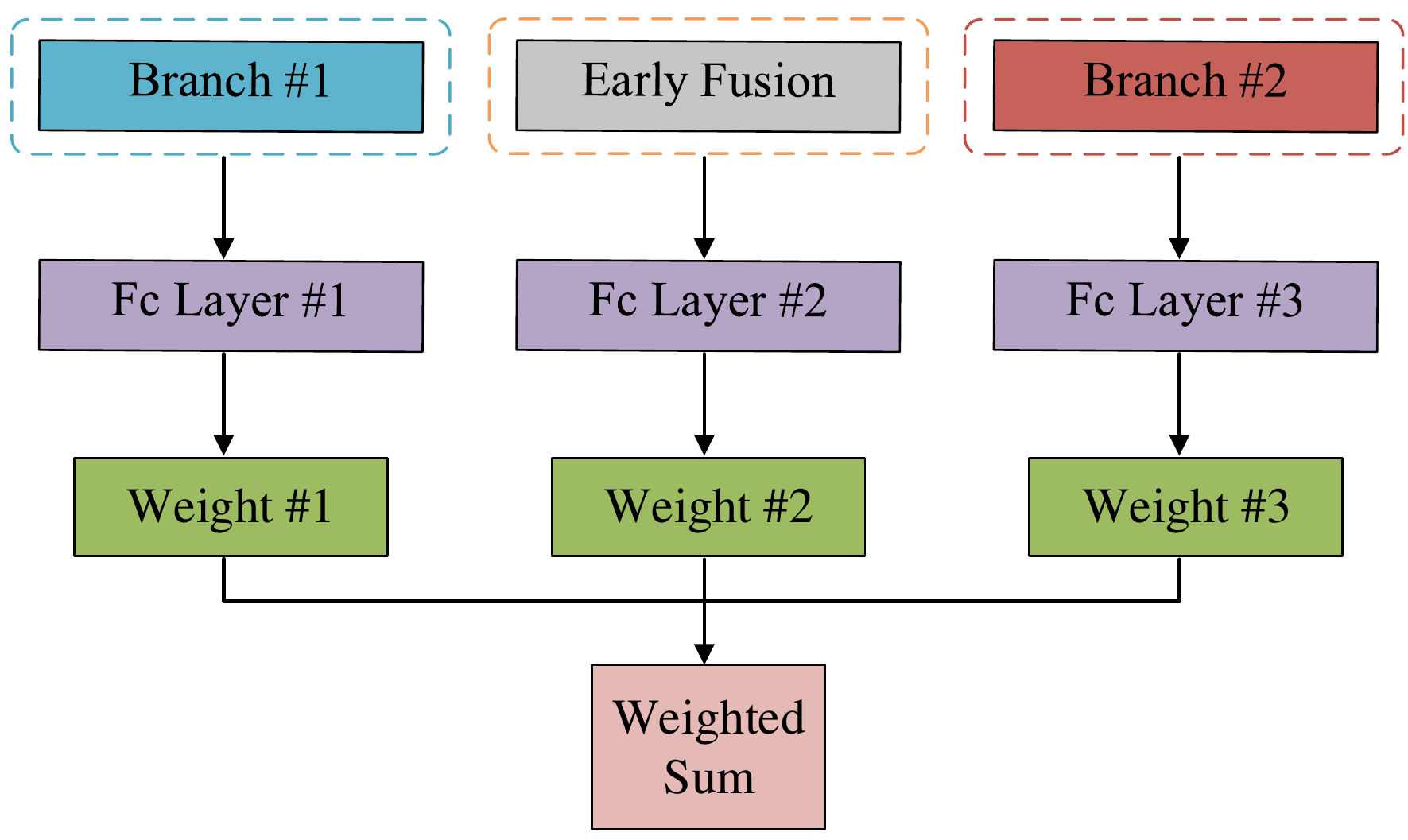}
\caption{The architecture of the weighted sum layer for three feature vectors.}\label{figfusion}
\end{centering}
\end{figure}

 Firstly, the main output which is more global to the side ones should be defined. In this work, the outputs of two branches and early fusion module are considered to be the side outputs of MCNN, and their output feature maps pass through the fully connected layers to get the corresponding feature vectors. A more global feature vector can be determined through a weighted sum operation shown in \autoref{figfusion} based on the three feature vectors, where trainable weights are used to mix the information and produce the main output. Therefore, four branches of output exist in the proposed architecture.
\par Next, advanced fusion mechanism is explored based on the deeply-supervised architectures. It can be seen that only one main classifier exist in \autoref{fig1.1}b, and the output of individual branches is indirectly utilized and optimized. Thus, further utilization of side outputs is adopted by adding extra classifiers. Adding classifiers to an architecture can be naturally associated with the idea of ensemble learning and joint decision making, which can prevent gradient vanishing, enhance the detail preservation and improve the interpretability of features \citep{inception,Lee2014Deeply}. While it was also used to implement multi-task learning \citep{Redmon2015You,Ren2015Faster,MEI201926} where the objective integrated multiple tasks through multiple metrics. Similar ideas also can be seen from the proposed architecture where three extra classifiers are added to provide complementary regularization and improve the performance. If we record the feature vectors obtained by phase branch, amplitude branch, early fusion module as $v_1$ to $v_3$ and weighted sum module as $v_m$, the main objective of the MCNN can be written as:
\begin{equation}
L_{main}=l_{e}(Y, v_m)=-Y\log(\sum_{i} w_i v_i)-(1-Y)\log(1-\sum_{i} w_i v_i),\label{8}
\end{equation}
where $Y$ is the label of ground truth, $w_i$ is the $i$th fusion weight in \autoref{figfusion} and $l_e$ is the cross entropy loss. The following objective come into being with the addition of three extra classifiers,
\begin{equation}
L_{side}=\sum_{i} \alpha_i l_{e}(Y,v_i)=\sum_{i} \alpha_i[-Y\log(v_i)-(1-Y)\log(1-v_i)].\label{7}
\end{equation}
\par Thus, the objective of the proposed architecture can be written as follows,
\begin{equation}
L_{obj}=L_{main}+L_{side}.\label{9}
\end{equation}
\par The prediction is the average of all the results of four classifiers:
\begin{equation}
Y_{pred}=Mean(Y_1,Y_2,Y_3,Y_{m}).\label{10}
\end{equation}
\par From the above representations, we can see that the proposed architecture has three extra classifiers, which correspond to the additional task of enhancing the presentation ability of the side outputs. Additional tasks are beneficial to optimize the network and present details, and all of them are used to improve the performance of the proposed architecture together.
\subsection{Depthwise separable convolutions based MCNN}
\label{sec:3.3}
\begin{figure}[!h]
\begin{centering}
\includegraphics[width=11.0cm,height=5.5cm]{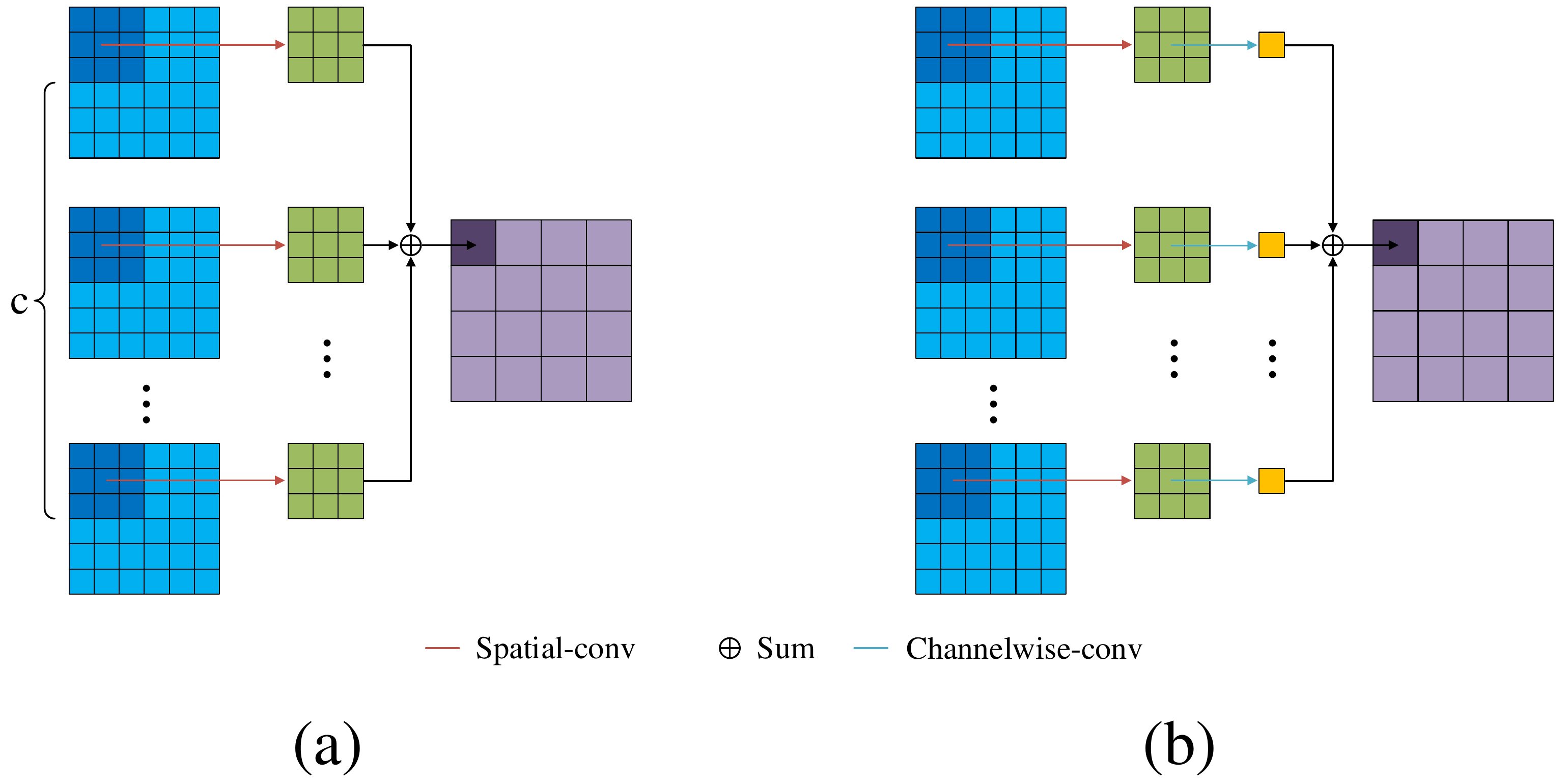}
\caption{Vanilla convolution and depthwise separable convolution with one group of kernel. (a) Vanilla convolution. (b) Depthwise separable convolution.}\label{figconv}
\end{centering}
\end{figure}

In this subsection, depthwise separable convolution \citep{xception} as an improved convolution, is used to better model the phase of PolSAR. Unlike optical images, the correlations between multi-channel phase information can express the structure information of the objects. Therefore, the problem is manifested in how to extract features from the data with certain correlations between channels. Notice that vanilla convolution acts on both spatial and channel dimensions as shown in \autoref{figconv}a, in which spatial filters and channelwise summation exist simultaneously. This means that the vanilla convolution kernels are responsible for extracting both spatial and channelwise information.
\par Depthwise separable convolution is based on the hypothesis that for the data which is closely related between channels, separating the vanilla convolution in spatial dimension and channel dimension may yield better results. It can be seen from \autoref{figconv}b that the channelwise summations in vanilla convolution are improved, and 1D-channelwise convolutions are added to the process. Thus, feature extraction processes are artificially separated from spatial dimension and channel dimension. When outputting multiple feature maps, vanilla convolution repeats spatial convolutions (green parts in \autoref{figconv}a) to produce multiple maps, while depthwise separable convolution repeats channelwise convolutions (yellow parts in \autoref{figconv}b). For input feature maps with the size of $w\times h\times c$, $K$ groups of $3\times3$ vanilla convolutional filters with $3\times3\times c\times K$ parameters are replaced by $c$ groups of $3\times3$ depthwise convolutions with $3\times3\times c$ parameters and $K$ groups of $1\times1$ pointwise convolutions with $1\times1\times c\times K$ parameters. Therefore, computational complexity is reduced due to the decrease in the size of kernels over multiple computations.
\par Compared with the vanilla convolution, depthwise separable convolution not only reduces the computational complexity, but also more suitable for the data with channel-correlations. To better model the phase information, vanilla convolutions in the phase branch of MCNN are replaced by depthwise separable convolutions, and the depthwise separable convolution based multi-task CNN (DMCNN) is built. Firstly, $c$ groups of $3\times3$ depthwise kernels are used to filter each channel of the maps spatially. Then, $K$ groups of $1\times 1\times c$ pointwise kernels are used to fuse the output of depthwise convolutions. Since the amplitude of the PolSAR is not significantly different from that of optical images, the architecture of the amplitude branch is preserved.
\section{Results}
\label{sec:4}
In this section, we verify the effectiveness of the proposed PolSAR image classification framework. Several CNNs-based classification algorithms are listed as objects of comparison. Experiment environment: PC with Intel Core i7-7700 CPU, Nvidia GTX-1060 GPU (6 GB memory), and 16 GB RAM. The deep learning platform \citep{tensorflow} is used to minimize the difficulty of algorithm implementation.
\subsection{Datasets description}
\label{sec:4.1}
We evaluate the proposed methods on three PolSAR benchmark datasets: AIRSAR Flevoland, ESAR Oberpfaffenhofen and EMISAR Foulum, which are commonly used in PolSAR image classification. Details of these datasets are listed as following:
\par
\begin{figure}
\begin{minipage}{0.32\linewidth}
  \centerline{\includegraphics[width=4.0cm,height=3.8cm]{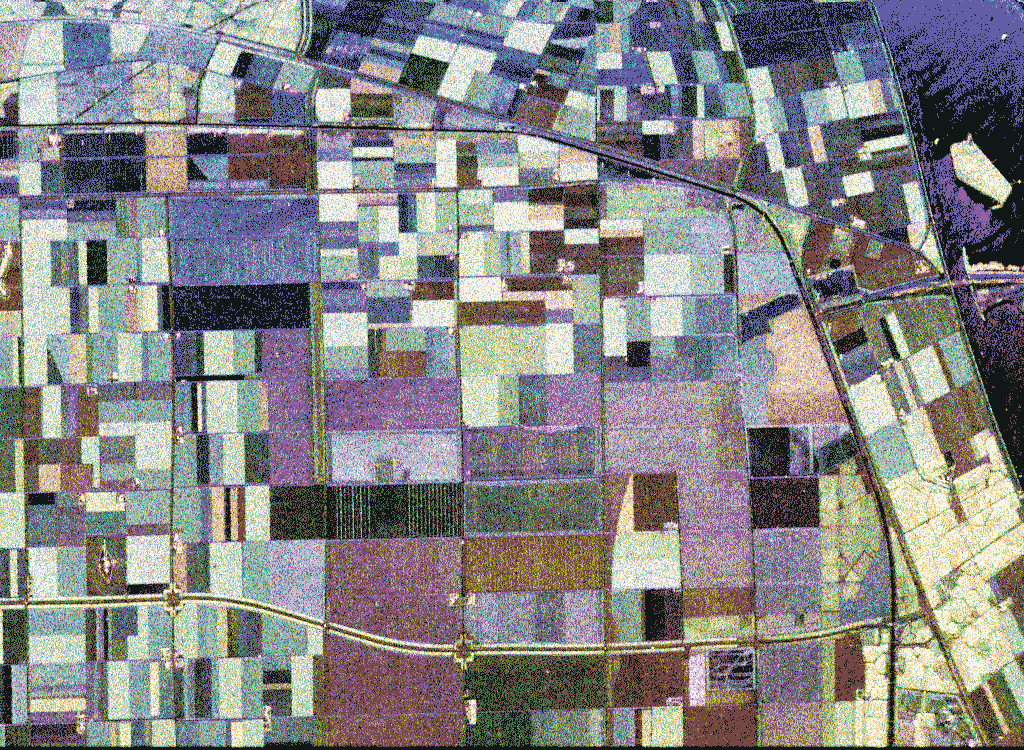}}
  \centerline{(a)}
\end{minipage}
\hfill
\begin{minipage}{0.32\linewidth}
  \centerline{\includegraphics[width=4.0cm,height=3.8cm]{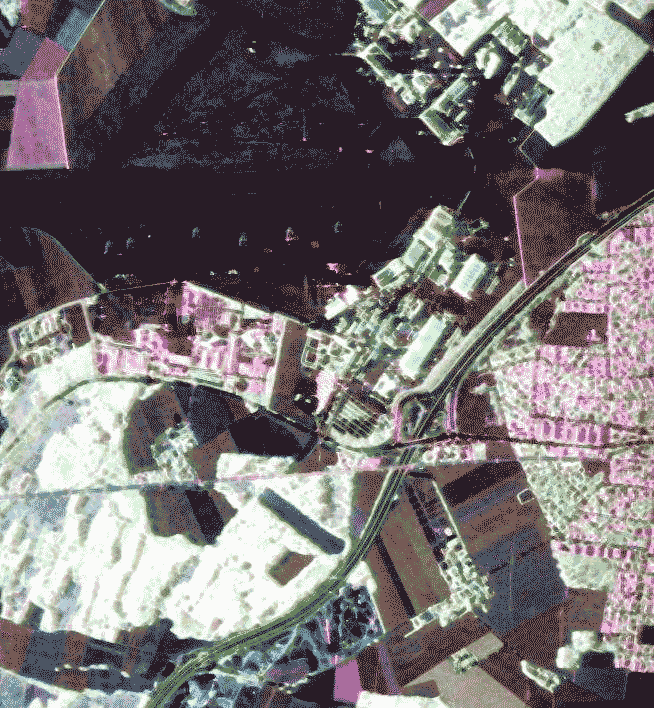}}
  \centerline{(b)}
\end{minipage}
\hfill
\begin{minipage}{0.32\linewidth}
  \centerline{\includegraphics[width=4.0cm,height=3.8cm]{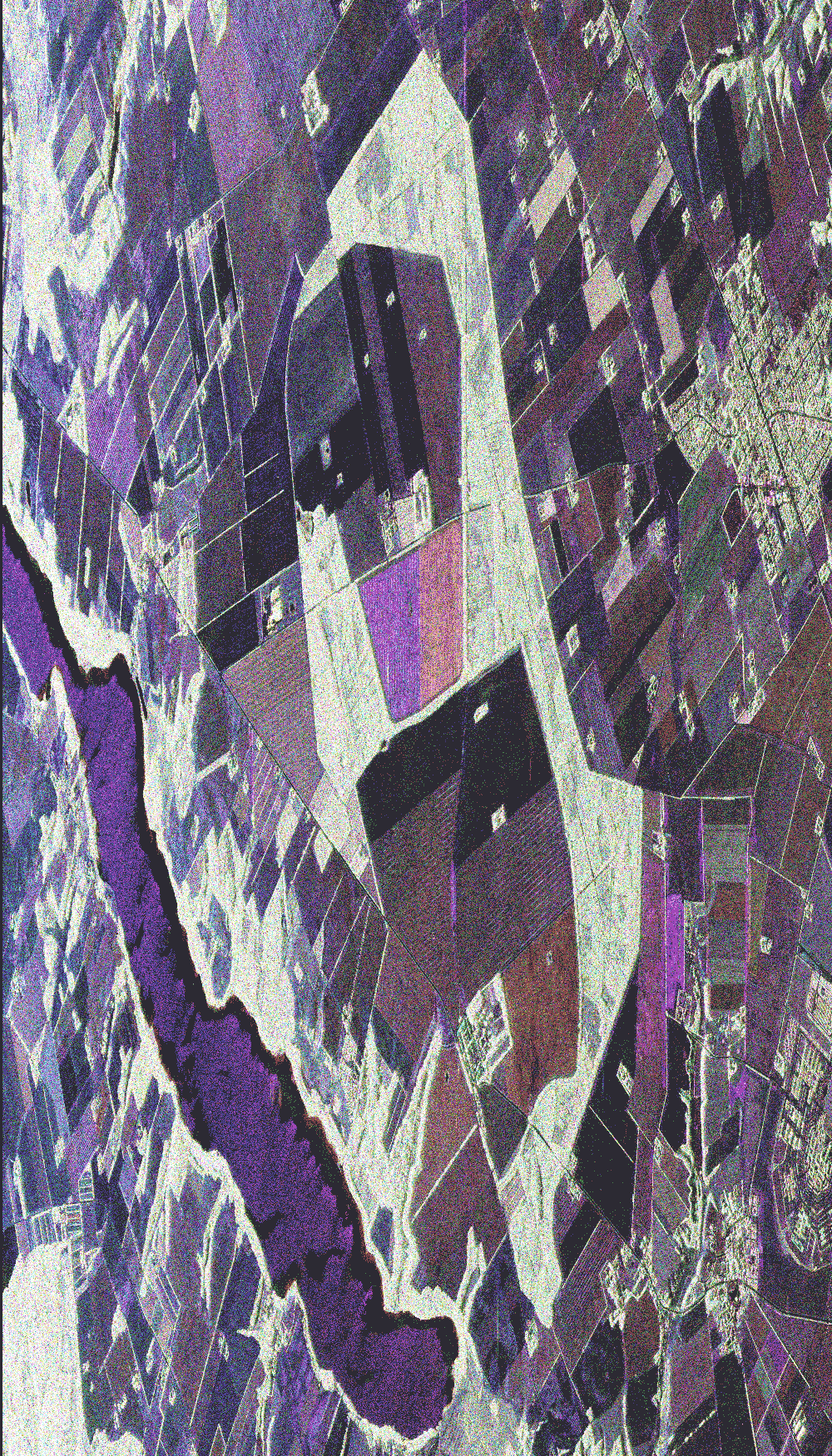}}
  \centerline{(c)}
\end{minipage}
\caption{Pauli image of the benchmarks. (a) AIRSAR Flevoland dataset. (b) ESAR Oberpfaffenhofen dataset. (c) EMISAR Foulum dataset.}
\label{figdatasets}
\end{figure}

 \textbf{AIRSAR Flevoland}: An L-band, full polarimetric image of the agricultural region of the Netherlands is obtained through NASA/Jet Propulsion Laboratory AIRSAR. The size of this image is $750\times1024$ and the spatial
resolution is $0.6m\times1.6m$. There are fifteen kinds of ground truth objects in \autoref{figdatasets}a including buildings, stem beans, rapeseed, beet, bare soil, forest, potatoes, peas, lucerne, barley, grass, water and three kinds of wheat. For the experiment of this map, 300 samples for each class are randomly selected as the training set.
\par \textbf{ESAR Oberpfaffenhofen}: An L-band, full polarimetric image of Oberpfaffenhofen, Germany, 1200$\times$1300 scene size, are obtained through ESAR airborne platform. Its Pauli color-coded image can be seen in \autoref{figdatasets}b. Each pixel in the map is divided into three categories: Built-up areas, wood land and open areas, except for some unknown regions. 600 labeled samples of each class are randomly selected as the training set.

\par \textbf{EMISAR Foulum}: The last full polarimetric image used in this experiment is the L-band image taken by EMISAR in Foulum, Denmark. EMISAR is a full polarized airborne SAR operating in L band and C band with resolution of $2m\times2m$ and mainly acquired and studied by Danish Center for Remote Sensing (DCRS). \autoref{figdatasets}c shows its Pauli RGB image. Five types of ground objects are marked in the image: River, rye, oats, winter wheat and coniferous. 200 labeled samples of each class are randomly selected as the training set.
\begin{table}
\centering
\caption{The detail information of three datasets used in experiments.}\label{table1}
\begin{tabular}{cccc}
\hline
Dataset     &Training num  &Testing num      \\
\hline
AIRSAR Flevoland	&4280 &66903\\
ESAR Oberpfaffenhofen	&1800 &113282	\\
EMISAR Foulum  &1000 &59905\\
\hline
\end{tabular}
\end{table}
\subsection{Experiment setup}
\label{sec:4.2}
To validate the significance of the proposed PolSAR image classification framework, the CNN architecture used in \citet{CNN2} and its VGGNet-style variant \citep{vgg} are chosen to be compared, which are noted as CNN and VGG for convenience. Experiments are implemented under different input forms to verify the validity of using amplitude and phase of PolSAR image instead of real parts and imaginary parts as input. CNN and VGG with the input of real and imaginary parts are abbreviated as CNN-v1 and VGG-v1, while the models using amplitude and phase input are noted as CNN-v2 and VGG-v2. The proposed multi-task CNN and the depthwise separable convolution based multi-task CNN only use the amplitude and phase as input, which are denoted as MCNN and DMCNN. During the training and testing of all the experiments, the PolSAR images are sliced into patches with the size of $14\times14$ and the stride of the sliding windows is set to be $1$.
\par In experiments, some mainstream designs are used to improve performance. After each convolution layer, batch normalization layer and ReLU layer are added. Each convolution layer and fully connected layer are followed by dropout layers, except for the last convolution layer in each convolution block \citep{ResNet,DenseNet}. Therefore, each layer becomes a series of cascaded combinations: Conv-Dropout-BN-ReLU in this paper. In order to speed up the optimization of objective function and obtain better approximate solution, adaptive moment estimation algorithm are used with the learning rate of 0.001 instead of ordinary stochastic gradient descent algorithm \citep{adam}. The size of all convolution kernels is $3\times3$ and the dropout rate is 0.5 for fully connected layers. For experiments on AIRSAR Flevoland and ESAR Oberpfaffenhofen, the ratio of dropout layers is 0.8 in convolution blocks and the numbers of convolution kernels are set as 32, 64, 64 for every convolution block in individual branches. For the densely connected fusion block, growth rate is set to be 16 and its first convolution layer outputs 4 times of growth rate feature maps. For the experiment on EMISAR Foulum, we do not abandon any data after the convolution layers. The numbers of convolution kernels are set as 12, 24, 24 for every convolution block in individual branches and the parameters in fusion block are changed to be 12 and 2 respectively. The multiplier factor of depthwise separable convolution is set to be 1.
\par To evaluate the performance of the involved algorithms, average accuracy (AA), overall accuracy (OA), kappa coefficient (Kappa) and $F_1$-score are chosen as criteria, which can be defined as follows \citep{input2}
\begin{equation}
AA=\frac{1}{c}\sum_{i=1}^c\frac{M_i}{N_i}, \; OA=\frac{\sum_{i=1}^c M_i}{\sum_{i=1}^c N_i},
\label{14}
\end{equation}
where $c$ is the number of categories. $M_i,N_i$ denote the number of $i$th class correctly classified samples and the number of the $i$th class labeled samples respectively.
\begin{equation}
Kappa=\frac{OA-P}{1-P}, \; with \; P=\frac{1}{N^2}\sum_{i=1}^cH(i,:)H(:,i),
\label{15}
\end{equation}
where $N$ is the number of testing samples. $H$ denotes the classification confusion matrix. The $F_1$-score of multi-class classification is calculated by averaging the $F_1$-score of each class. $F_1$-score of the $i$th class can be obtained as follows,
\begin{equation}
{F_1}^i=\frac{2TP_i}{2TP_i+FP_i+FN_i},
\label{16}
\end{equation}
where $TP$ represents the number of correct positive samples, $FN$ and $FP$ represent the number of mistaking other classes for $i$th class and wrong prediction of $i$th class respectively. Thus, $F_1$-score can be calculated by
\begin{equation}
F_1=(\frac{1}{c}\sum_{i=1}^c {F_1}^i)^2.
\label{17}
\end{equation}
\subsection{Comparison of results on benchmarks}
\label{sec:4.3}
Experiments are carried out on the benchmark datasets based on the above settings. The following results and analyses can be obtained:
\begin{enumerate}[(1)]
  \item \textbf{Results on AIRSAR Flevoland dataset}: Classification results of the whole map obtained by different methods can be seen from \autoref{figresult1}. The proposed methods, especially DMCNN, achieve higher completeness of the terrains in the classification maps. As shown in \autoref{fig_expadd1}, the proposed DMCNN and MCNN converge to high precisions after several iterations. This observation proves that the proposed framework can achieve good classification results on AIRSAR dataset. From the experimental results shown in \autoref{tableAIRSARcompare}, it can be seen that the proposed methods achieve the best and the second results respectively. PolSAR tailored architectures have better performance than the one used in optical image processing. This result confirms the importance of designing tailored architectures for the task of PolSAR image classification. Moreover, it can be seen that the classification accuracy of DMCNN is higher than the other methods. This means that depthwise separable convolution can improve the effects of extracting information from the phase of PolSAR images. The cause of this phenomenon may be the potential correlations lying in the phase.

\begin{sidewaystable}
\tiny
\centering
\caption{Comparison of experimental results $(\%)$ on AIRSAR Flevoland dataset. For convenience, C1 to C15 refer to fifteen different categories: Buildings, rapeseed, beet, stem beans, peas, forest, lucerne, potatoes, bare soil, grass, barley, water, wheat one, wheat two and wheat three.}\label{tableAIRSARcompare}
\begin{tabular}{cccccccccccccccccc}
\hline
Method     &C1  &C2 &C3 &C4 &C5 &C6  &C7 &C8 &C9 &C10 &C11 &C12 &C13 &C14 &C15 &AA &OA   \\
\hline
CNN-v1	&100.00      &84.62  &99.91&100.00    &100.00       &99.96
 &96.22 &99.47 &99.39  &93.75     &99.78  &93.95
 &97.14 &97.43 &93.72 &97.02 &96.13\\
VGG-v1	&100.00        &94.87 &100.00       &100.00       &98.81  &100.00
 &99.18   &98.12  &100.00    &100.00       &97.37  &75.78
 &90.66  &94.51   &99.91 &96.61 &93.61\\
CNN-v2  &100.00         &96.14 &100.00    &100.00         &98.47  &99.93
 &99.20  &99.53 &100.00       &97.58  &67.27  &98.22
 &87.55  &89.49    &91.49 &94.99 &92.73\\
VGG-v2  &100.00     &98.65 &100.00     &100.00       &65.60  &99.96
 &99.79  &96.17 &100.00  &100.00      &99.90 &98.92
 &98.37 &98.83 &97.12  &96.89 &96.52\\
MCNN  &100.00 &100.00   &100.00  &100.00 &100.00   &95.34 &99.77 &100.00 &100.00 &100.00  &98.15 &98.96 &92.69   &99.13  &99.92  &98.93 &98.09\\
DMCNN  &100.00         &99.92 &100.00        &100.00      &100.00         &99.58
 &100.00        &99.91  &100.00    &100.00        &99.98 &100.00
 &93.67 &100.00        &99.69  &\textbf{99.52} &\textbf{98.77}\\
\hline
\end{tabular}
\end{sidewaystable}

\begin{figure}[!h]
\begin{centering}
\noindent\makebox[\textwidth][c] {
\includegraphics[width=0.90\paperwidth, height=11.5cm]{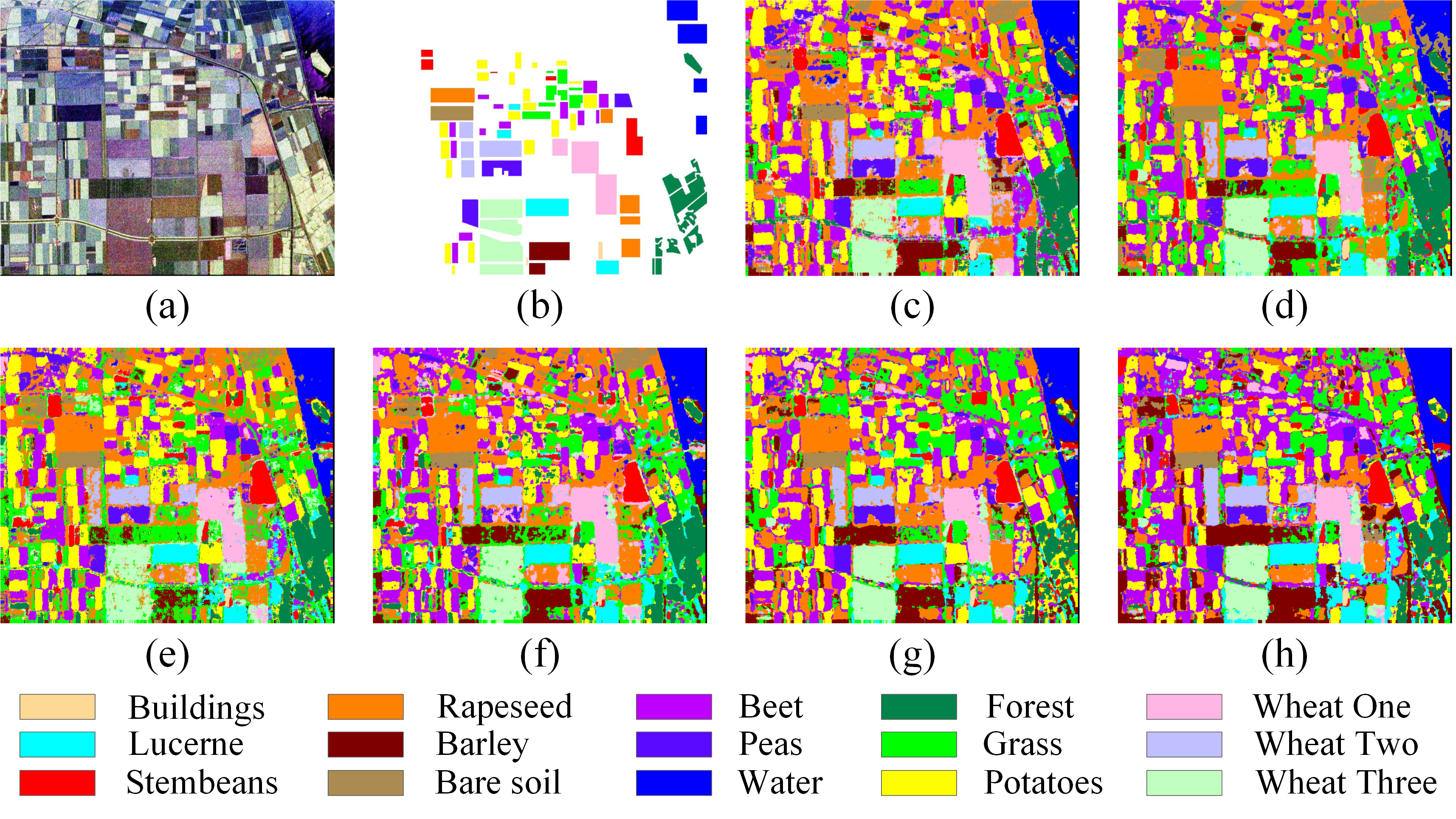}}
\caption{Classification results of whole map on AIRSAR Flevoland data with different methods. (a) Pauli RGB map. (b) Ground truth map. (c) CNN-v1. (d) VGG-v1. (e) CNN-v2. (f) VGG-v2. (g) MCNN. (h) DMCNN.}\label{figresult1}
\end{centering}
\end{figure}

\begin{figure}
\begin{minipage}{0.48\linewidth}
  \centerline{\includegraphics[width=6.6cm,height=4.2cm]{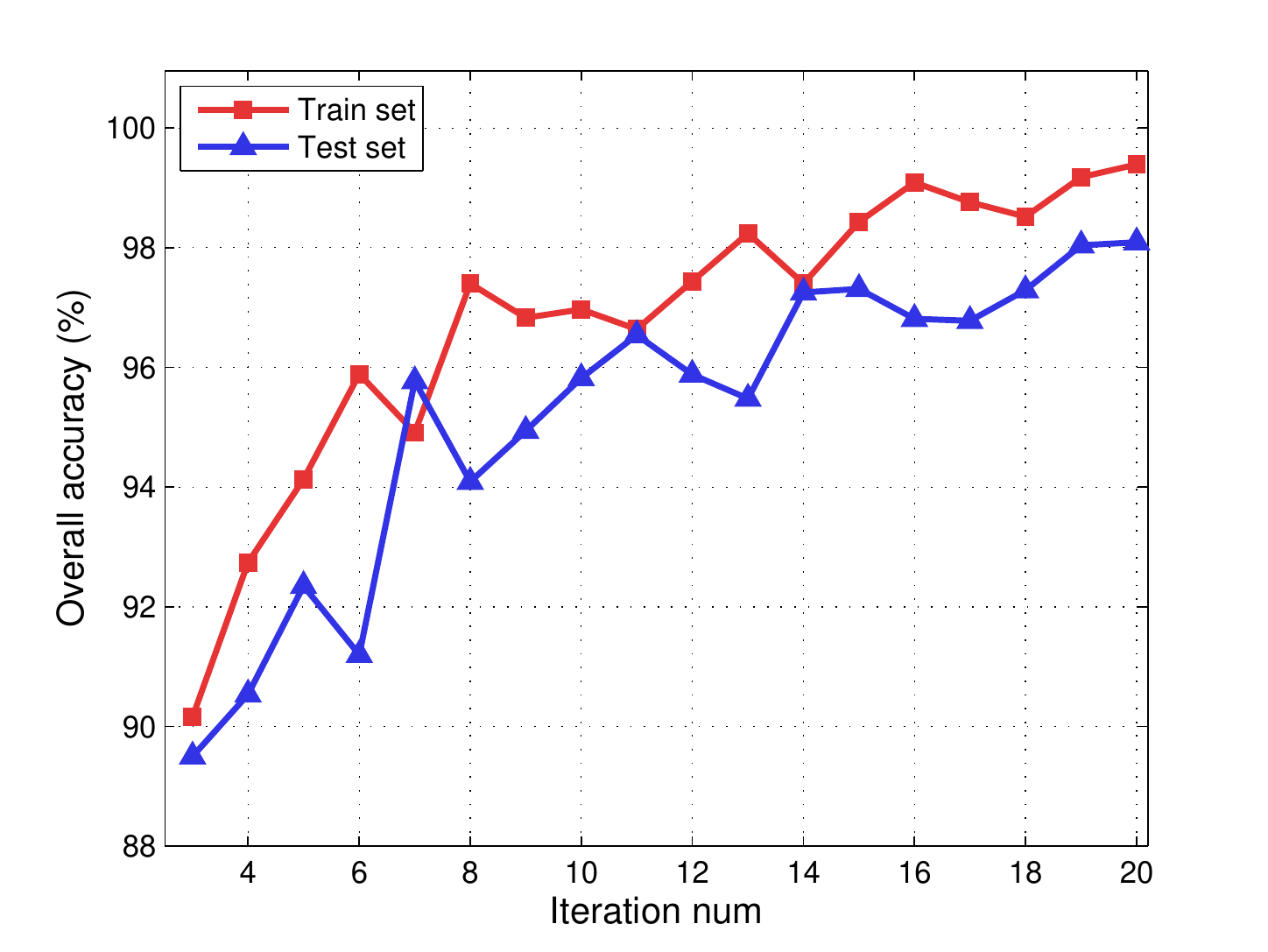}}
  \centerline{(a)}
\end{minipage}
\hfill
\begin{minipage}{0.48\linewidth}
  \centerline{\includegraphics[width=6.6cm,height=4.2cm]{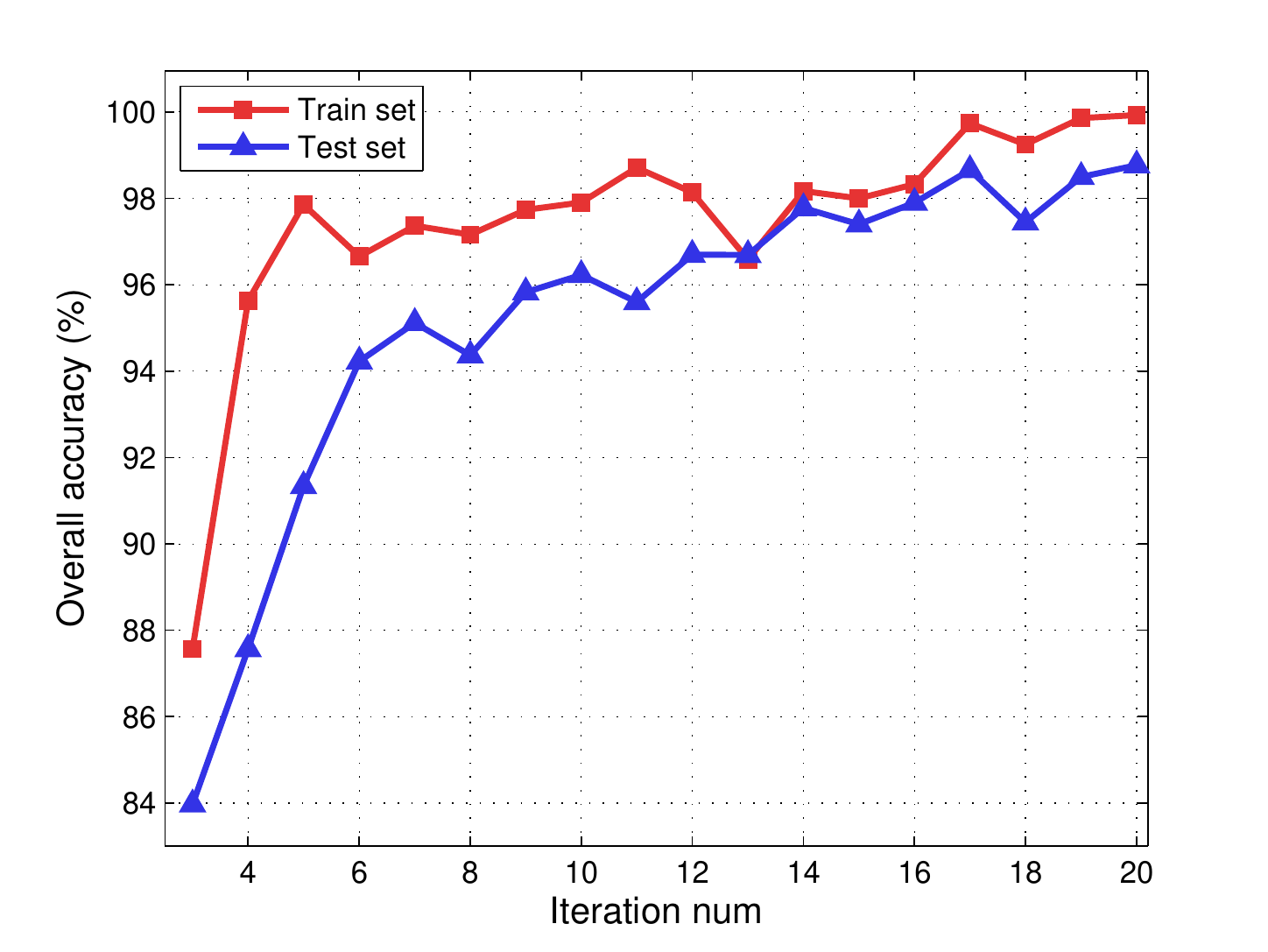}}
  \centerline{(b)}
\end{minipage}
\caption{Overall accuracy curves of the proposed methods on the train and test sets of AIRSAR Flevoland dataset respectively. (a) MCNN. (b) DMCNN.}
\label{fig_expadd1}
\end{figure}

  \item \textbf{Result on ESAR Oberpfaffenhofen}: The whole map experimental results are presented in \autoref{figresult2}, from which we can see that the proposed method can better distinguish between built-up areas and wood land. The accuracy curves of the proposed methods on ESAR Oberpfaffenhofen dataset are listed in \autoref{fig_expadd2}, from which the highest overall accuracy reaches 98.06$\%$ on the test set. \autoref{tableESARcompare} shows the experimental results of each algorithm on Oberpfaffenhofen dataset. It can be seen that for the same architectures (CNN and VGG), the classification results obtained by the input of amplitude and phase are better than the one of real and imaginary, and the change of input form almost improve the accuracy of every class. This observation shows that the change of input form may increase the classification accuracy without changing the classifiers. Besides, the classification accuracy of the proposed methods are significantly improved compared with the contrast models, which is consistent with the theoretical guidance of representation learning.
  \begin{table}
\centering
\caption{Comparison of experimental results $(\%)$ on ESAR Oberpfaffenhofen dataset. For convenience, C1 to C3 refer to different categories: Built-up areas, wood land and open areas.
}\label{tableESARcompare}
\begin{tabular}{cccccc}
\hline
Method     &C1  &C2 &C3  &AA &OA   \\
\hline
CNN-v1	&82.09 &91.17 &99.09 &90.78 &89.71 \\
VGG-v1	&63.98  &99.87 &97.77 &87.21 &84.85\\
CNN-v2  &90.44 &96.79  &96.87 &95.03 &94.26\\
VGG-v2  &85.32&99.17 &99.48 &94.66 &93.69\\
MCNN  &92.99 &99.02 &99.98 &97.33 &96.86\\
DMCNN  &95.23 &99.93 &99.98 &\textbf{98.38} &\textbf{98.06}\\
\hline
\end{tabular}
\end{table}

\begin{figure}[!h]
\begin{centering}
\noindent\makebox[\textwidth][c] {
\includegraphics[width=0.80\paperwidth, height=12.5cm]{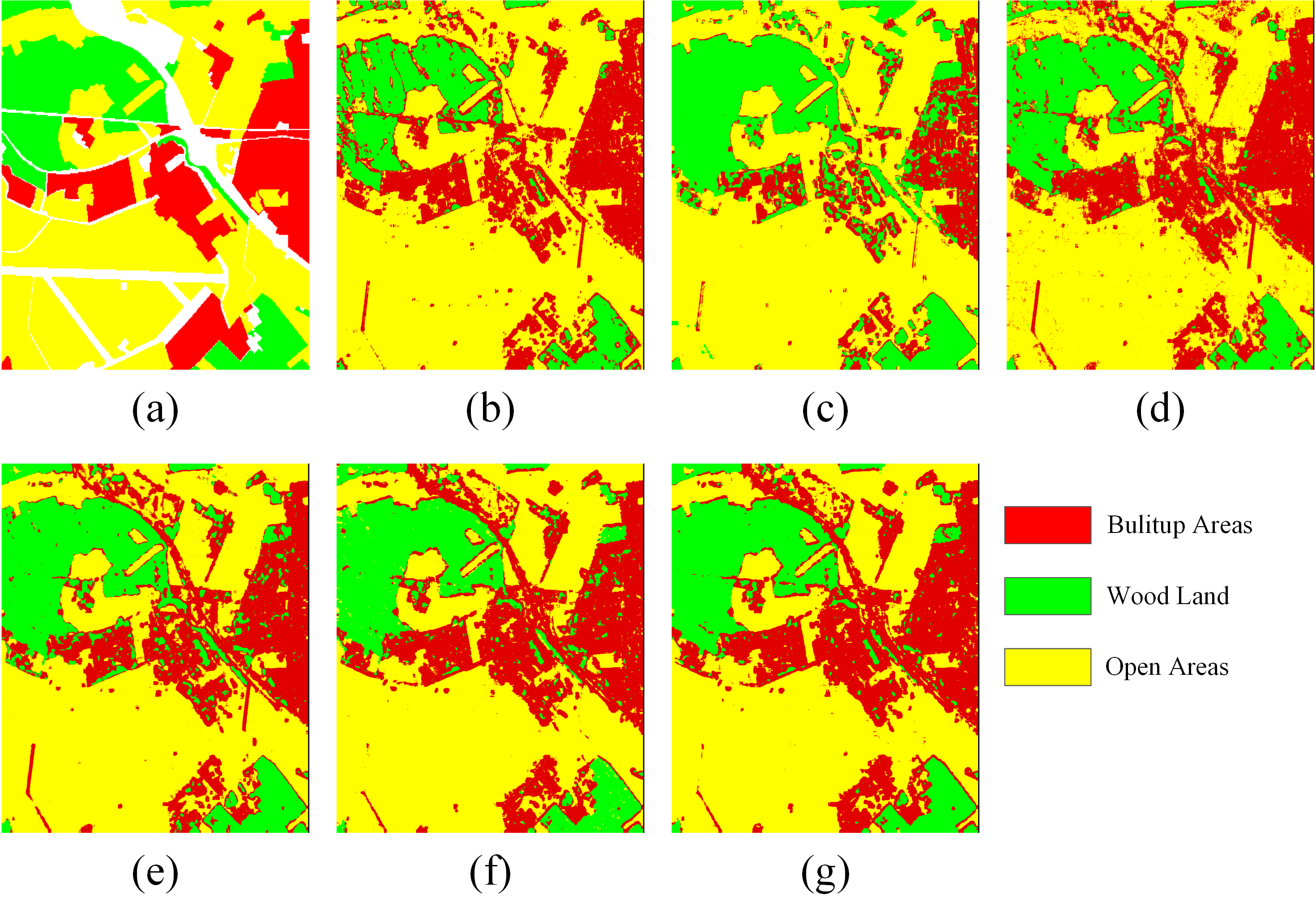}}
\caption{Classification results of whole map on ESAR Oberpfaffenhofen data with different methods. (a) Ground truth map. (b) CNN-v1. (c) VGG-v1. (d) CNN-v2. (e) VGG-v2. (f) MCNN. (g) DMCNN.}\label{figresult2}
\end{centering}
\end{figure}

\begin{figure}
\begin{minipage}{0.48\linewidth}
  \centerline{\includegraphics[width=6.6cm,height=4.2cm]{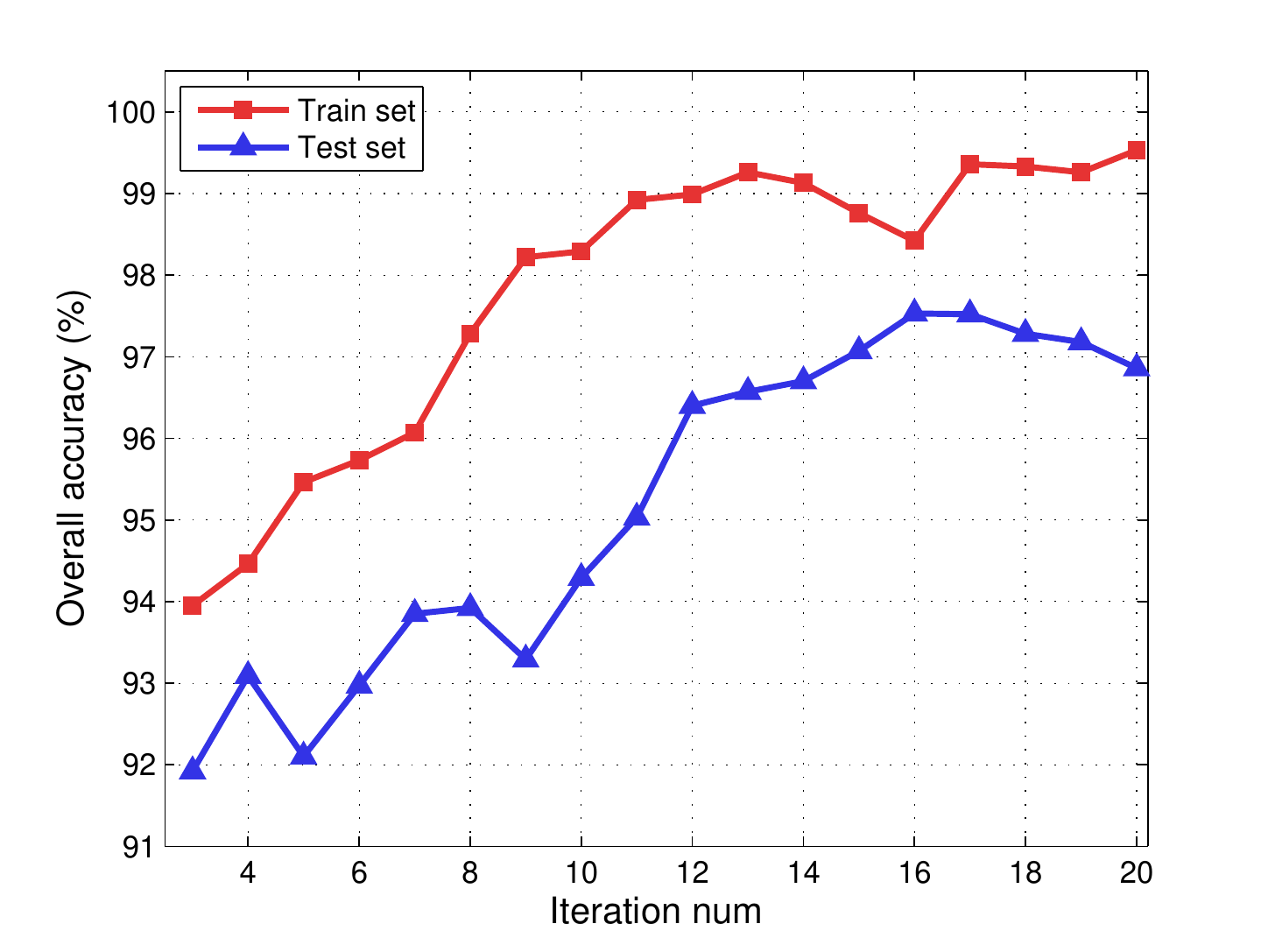}}
  \centerline{(a)}
\end{minipage}
\hfill
\begin{minipage}{0.48\linewidth}
  \centerline{\includegraphics[width=6.6cm,height=4.2cm]{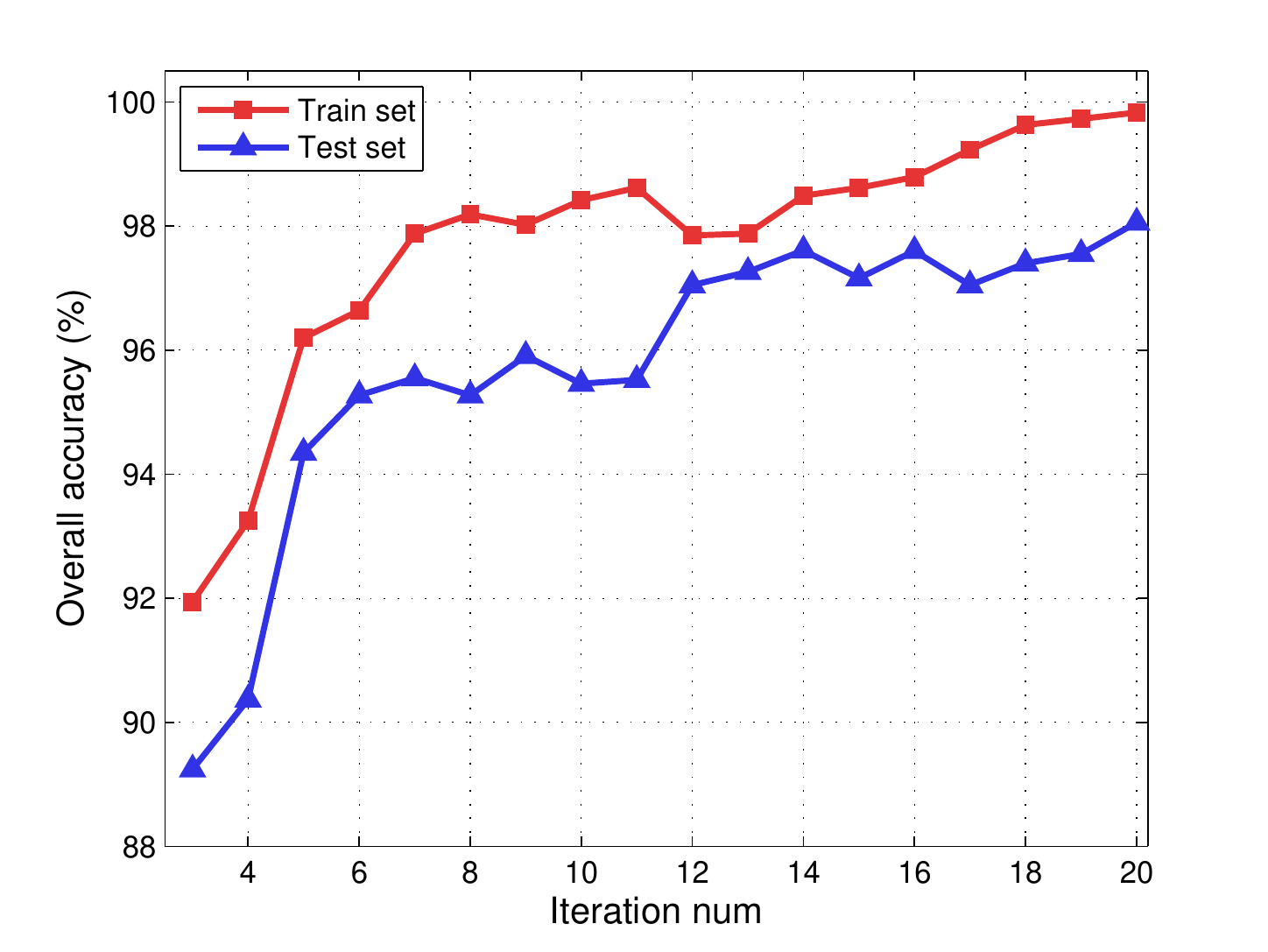}}
  \centerline{(b)}
\end{minipage}
\caption{Overall accuracy curves of the proposed methods on the train and test sets of ESAR Oberpfaffenhofen dataset respectively. (a) MCNN. (b) DMCNN.}
\label{fig_expadd2}
\end{figure}

  \item \textbf{Result on EMISAR Foulum}: Results of the experiments on EMISAR Foulum dataset can be seen from \autoref{tableEMISARcompare}. It can be seen that the proposed methods still achieve good results. Unlike the previous experiments, the change of input results in a decrease in the accuracy of CNN and VGG. This observation shows that an architecture is not suitable for all data forms and it is necessary to adjust the network architectures to fit different kinds of inputs. As the results shown in \autoref{figresult3}, the edge of the river is difficult to identify affected by the angle of view. DMCNN has the best recognition performance for rivers, which benefits from the utilization of multi-view information of PolSAR. MCNN has better classification results for the other four types of terrains than comparison methods. The training and testing accuracy of the proposed methods which converge to relatively high values after iterations are shown in \autoref{fig_expadd3}. It can be seen that the differences between the proposed models are not obvious.
  \begin{table*}
\centering
\caption{Comparison of experimental results $(\%)$ on EMISAR Foulum dataset. For convenience, C1 to C5 refer to five different categories: River, rye, oats, winter wheat and coniferous.
}\label{tableEMISARcompare}
\begin{tabular}{cccccccc}
\hline
Method     &C1  &C2 &C3 &C4 &C5  &AA &OA   \\
\hline
CNN-v1	&85.16 &89.82 &100.00     &99.37  &99.64 &94.80 &94.68\\
VGG-v1	&78.60  &98.75&100.00   &100.00  &99.45&95.36 &93.51\\
CNN-v2  &83.68  &66.09 &96.05 &99.69  &99.99 &89.10 &92.27\\
VGG-v2  &70.85 &69.79 &96.50  &98.28  &99.89 &87.06 &88.91\\
MCNN  &98.98 &99.92 &100.00 &100.00 &100.00&\textbf{99.78} &99.71\\
DMCNN  &99.84 &98.79 &99.32 &100.00 &100.00&99.59 &\textbf{99.83}\\
\hline
\end{tabular}
\end{table*}

\begin{figure}[!h]
\begin{centering}
\noindent\makebox[\textwidth][c] {
\includegraphics[width=0.80\paperwidth, height=14.0cm]{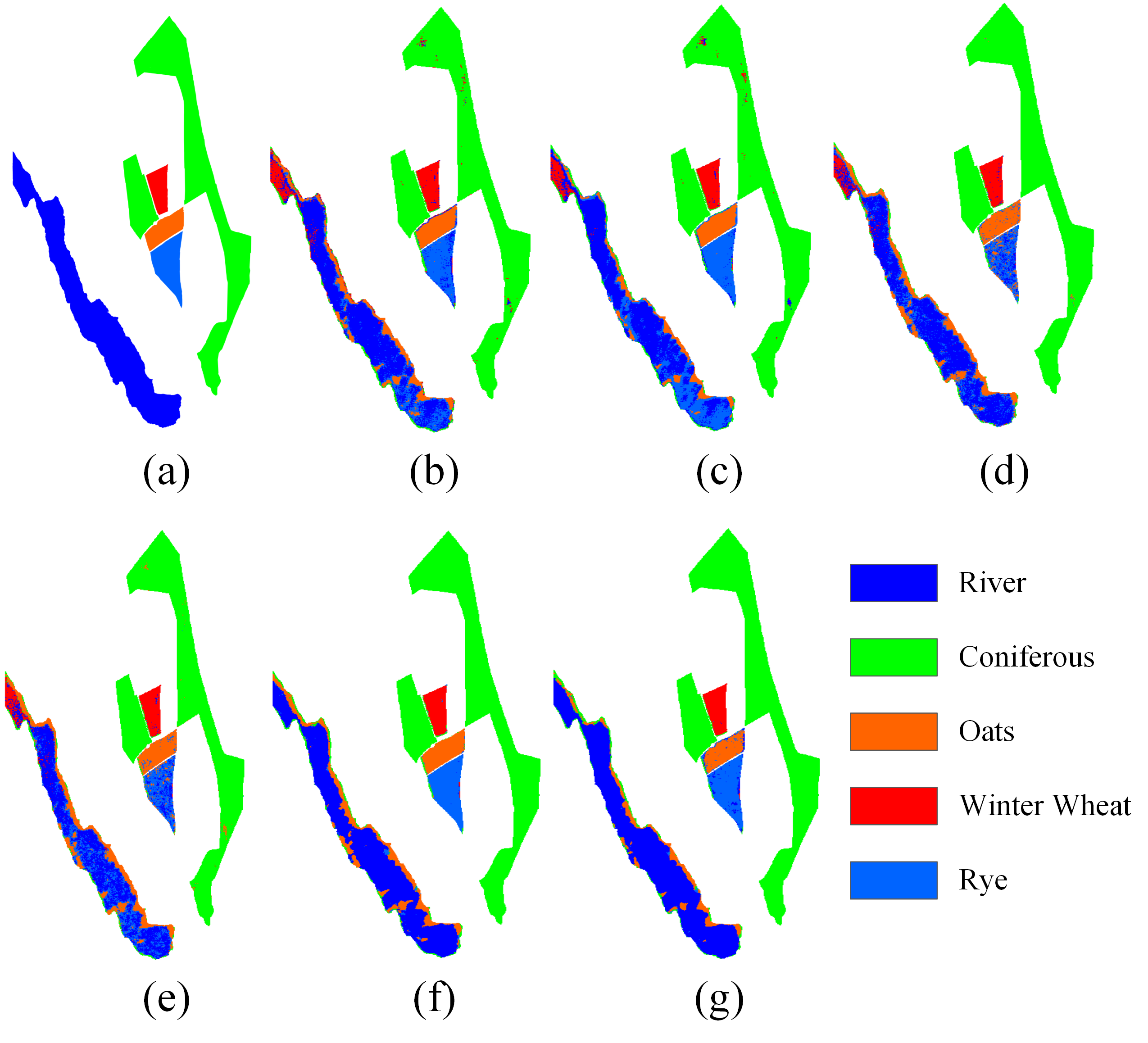}}
\caption{Classification results overlaid with the ground truth map on EMISAR Foulum data with different methods. (a) Ground truth map. (b) CNN-v1. (c) VGG-v1. (d) CNN-v2. (e) VGG-v2. (f) MCNN. (g) DMCNN.}\label{figresult3}
\end{centering}
\end{figure}
\end{enumerate}

\begin{figure}
\begin{minipage}{0.48\linewidth}
  \centerline{\includegraphics[width=6.6cm,height=4.2cm]{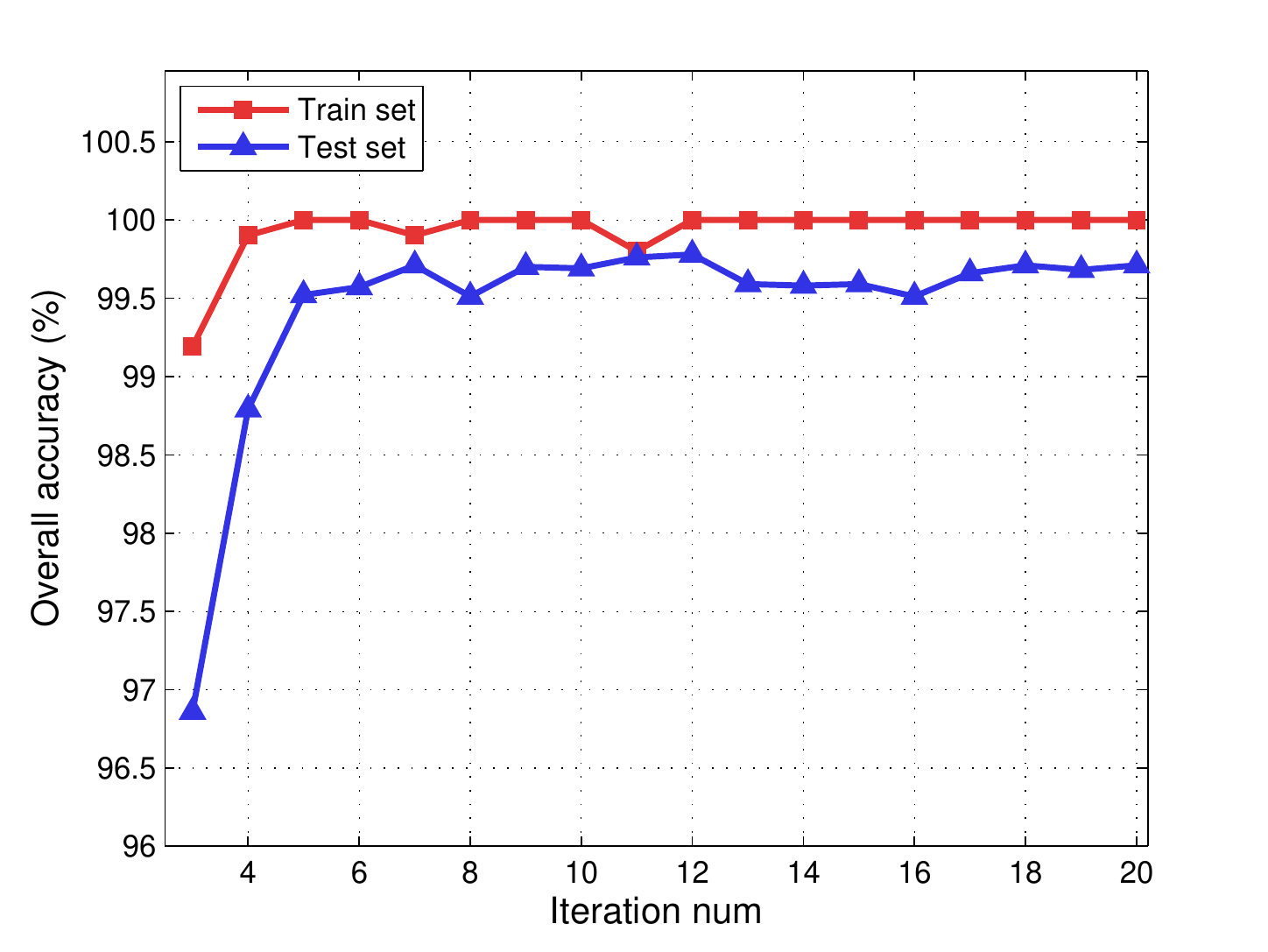}}
  \centerline{(a)}
\end{minipage}
\hfill
\begin{minipage}{0.48\linewidth}
  \centerline{\includegraphics[width=6.6cm,height=4.2cm]{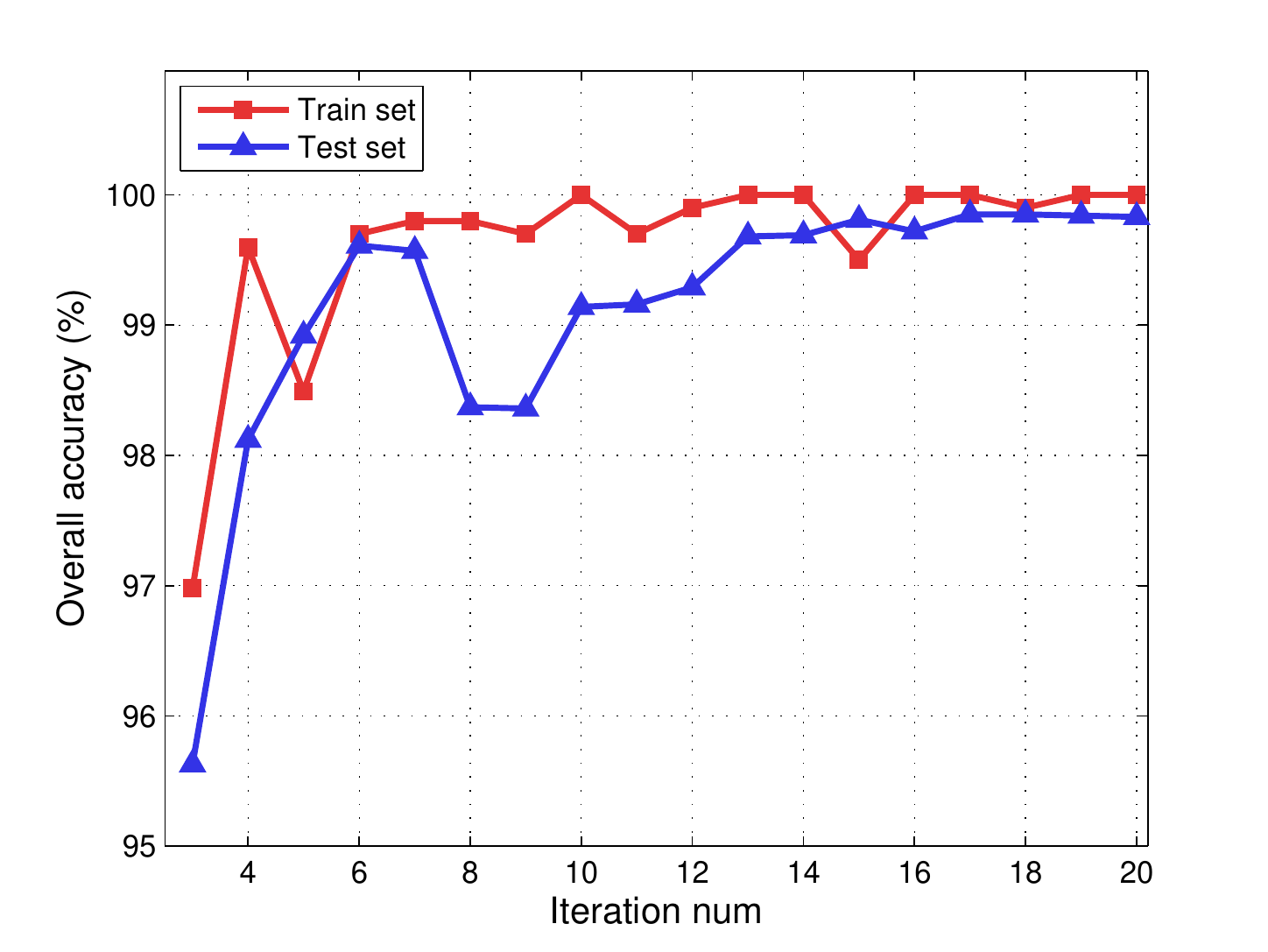}}
  \centerline{(b)}
\end{minipage}
\caption{Overall accuracy curves of the proposed methods on the train and test sets of EMISAR Foulum dataset respectively. (a) MCNN. (b) DMCNN.}
\label{fig_expadd3}
\end{figure}

A summary of the above experimental results can be given as follows,
\begin{itemize}
  \item The proposed classification methods can improve the classification accuracy on each benchmark dataset. The reason is closely related to the transformation of the input form. The experimental results prove that using amplitude and phase of PolSAR data as the input of CNNs has played an important role in promoting classification performance. One possible reason may be the significant difference existing between PolSAR images and optical images, so following the pattern of optical image processing will lose the unique information of PolSAR images, and the information of complex-valued PolSAR data can be better preserved by the form of amplitude and phase than real and imaginary parts.
  \item Well-designed network architectures according to the characteristics of input data are important for improving the classification accuracy. In the experiments, the optical image classification used CNNs are not as effective as the proposed PolSAR tailored architectures when the amplitude and phase are extracted as the input. Among them, the two-stream architecture helps a lot in enhancing the performance. Different types of information can be differently and pertinently processed with the help of multiple network branches. This reflects the importance of adjusting architectures according to the input data when applying deep learning methods to PolSAR image classification.
  \item Depthwise separable convolution can better model the PolSAR phase information. DMCNN has shown strong competitiveness in many groups of experiments. It is well known that underlying correlations which are helpful to recognize the ground target exist in the phase of PolSAR images. Thus, how to excavate the correlations should be considered during the network modeling. Compared with 3D convolution with huge computational burden, depthwise separable convolution provides a new way to mine the information contained in the phase.
\end{itemize}

\subsection{Ablation experiments}
\label{sec:4.4}
As shown in \autoref{fig4}, the proposed PolSAR classification framework can be decomposed into multiple segments. Each of the components is designed by hand and expert knowledge is heuristically incorporated. Thus, the truth and validity of each component should be empirical proved by ablation experiments. In this subsection, ablation experiments are carried out to verify the rationality of the proposed framework. Six architectures are decomposed to observe the effectiveness of the addition of advanced modeling tricks. They are simply recorded as $M1$ to $M6$ in the order of appearance for convenience.
\par\textbf{Two-stream CNN without fusion ($M1$)}: A two-stream CNN model \citep{twostream1} dealing with the amplitude and phase separately, shown in \autoref{fig1.1}a, is constructed to verify the significance of feature fusion. Among them, two branches share parameters. Softmax classifiers are used at the end of each branch to get the classification probability. The output of individual branches is obtained by processing the amplitude and phase separately, and the average of the two output probabilities are used as the final result.
\par\textbf{Two-stream CNN with convolution fusion ($M2$)}: The overview of this architecture can be seen from \autoref{fig1.1}b. Compare to $M1$, the classifiers are replaced and a vanilla convolution layer is added behind the convergence of two branches. Such a simple module is often used to fuse the information obtained from the two branches and explore higher-level features \citep{twostream2}. In this way, information interaction exists between the two branches of the architecture. Fully connected layers and softmax classifier are used to classify the output of fusion module.
\par\textbf{Densely connected fusion ($M3$)}: A densely connected convolution block \citep{DenseNet} is used for information fusion instead of a simple convolution layer in the proposed framework. In theory, such a fusion module can store more high-order information. To verify the effectiveness of this trick, the fusion mechanism in $M2$ is replaced by the proposed fusion module.
\par\textbf{Adding extra classifiers ($M4$)}: The classifiers in $M1$ are enabled again and added to $M3$ to build this architecture. Thus, the idea of deeply-supervised architectures is introduced to prevent the gradient vanish caused by the deepening of the network. The output of the side classifiers are also added to the final decision \citep{Lee2014Deeply,inception}.
\par\textbf{Multi-task CNN ($M5$)}: This part is the proposed MCNN. Compared with $M4$, deeply-supervised skip connections, shown in \autoref{fig2.4}, is added to make better use of the existing information for comprehensive decision making.
\par\textbf{Depthwise separable convolution based model ($M6$)}: Depthwise separable convolution is introduced in order to make full use of the unique phase information of the PolSAR image, which has been extensively described before.
\begin{figure}
  \centering
  \includegraphics[width=8.0cm,height=6.0cm]{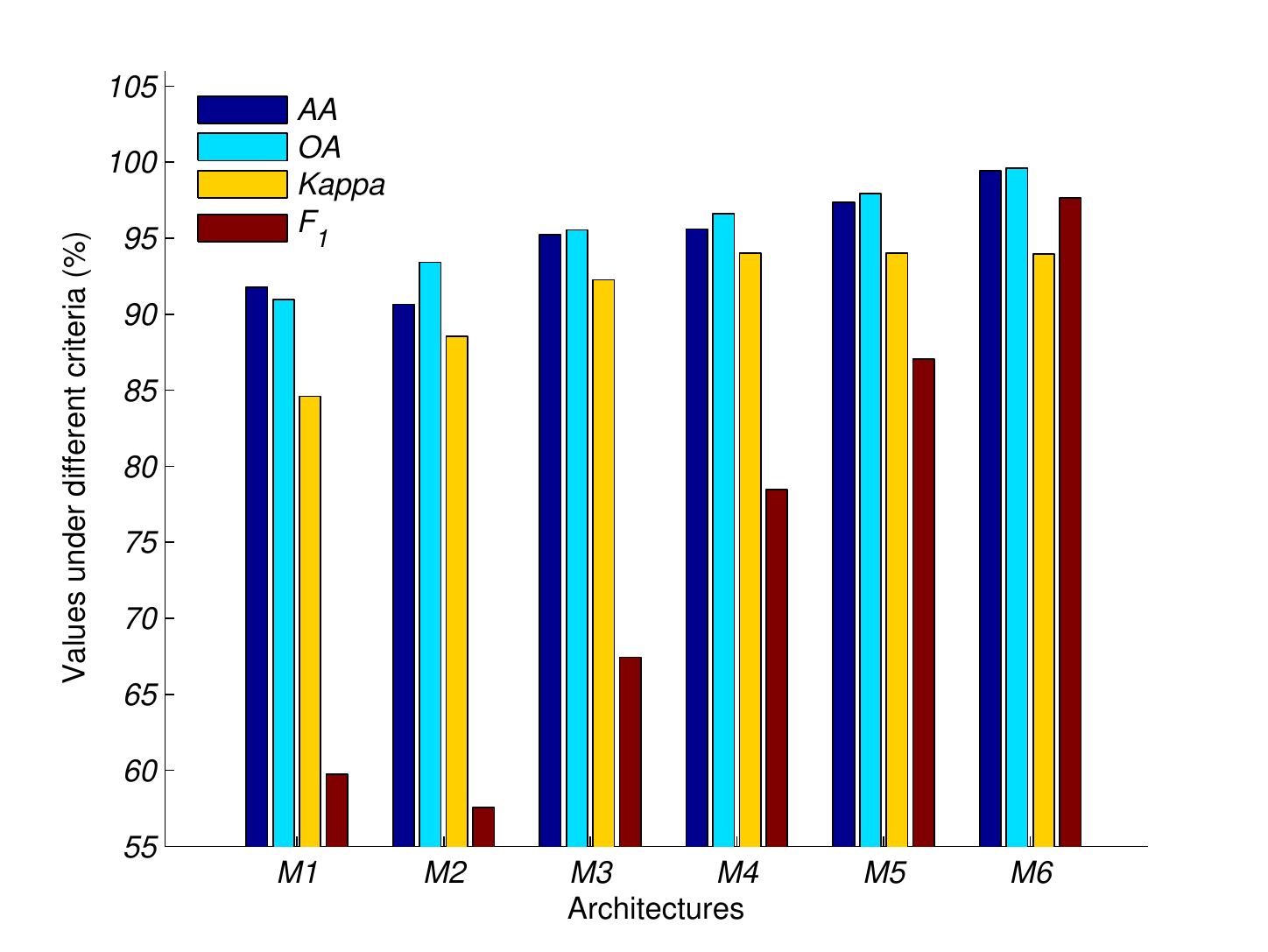}\\
  \caption{The classification accuracy results of decomposed architectures of the proposed framework on EMISAR dataset.}\label{figcompare}
\end{figure}
\par The results are shown in \autoref{figcompare}, it can be seen that the accuracy of the architectures $M1$ to $M6$ is gradually increasing under different criteria. Carefully speaking, it can be found by comparing the results of $M1$ to $M3$ that the CNN architecture with the feature fusion module has higher precision and the densely connected fusion mechanism has stronger ability than the convolution fusion. The result of $M4$ is better than the previous three architectures, which shows that the addition of side classifiers is beneficial to the network optimization and improvement of accuracy. Moreover, the performance of $M4$ is worse than $M5$, which shows that the addition of advanced fusion layer improves the accuracy of classification. The comparison between $M5$ and $M6$ further confirms the previous conclusion that phase information can be better used to obtain more accurate classification results.
\section{Discussion}
\label{sec:add}
In this section, intuitive comparisons are given between the proposed framework and other common methods. Firstly, differences between the proposed method and the widely used deep learning based PolSAR classification methods are discussed. Then, the proposed MCNN is contrasted with some common architectures to show its innovations.
\par Compared to other deep learning based PolSAR classification methods, the main characteristics of the proposed framework can be summarized as follows,
\begin{itemize}
  \item A novel input form of PolSAR data for CNNs is explored for the first time, which keeps the integrity of the original data and avoids complex-valued operations.
  \item Tailored CNN architectures are proposed according to the used input form for PolSAR classification. More network branches are added to process the different kinds of input information respectively. A densely connected convolutional fusion module is used for preliminary information interaction.
  \item Extra classifiers are introduced based on the concept of multi-task learning and deeply-supervised architectures. This measure not only makes the optimization more convenient, but also enhances the performance by fusing multi-level outputs.
\end{itemize}
\par
\begin{figure}[h]
\begin{centering}
\includegraphics[width=10.0cm,height=7.0cm]{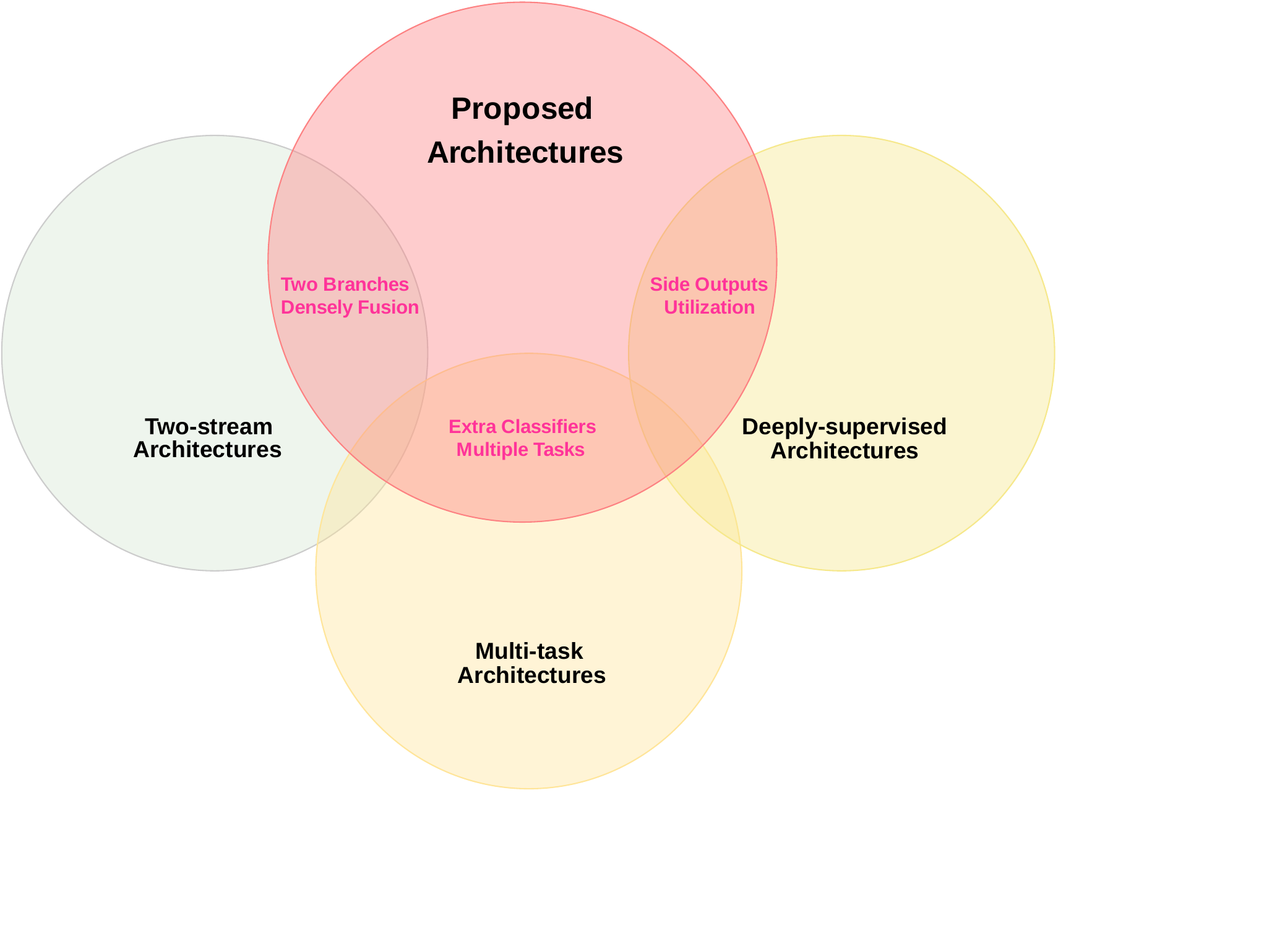}
\caption{Illustration of the relationship between the proposed architecture and related ones. All sets in the Venn diagram are subsets of CNNs.}\label{fig_venn}
\end{centering}
\end{figure}

 The improvements are carried out from the following two aspects in this paper. Firstly, amplitude and phase of complex-valued PolSAR data are extracted as the input of CNNs-based PolSAR image classification methods, which can be seen as the greatest highlights of this work. Furthermore, a tailored MCNN is studied to match the improved input form on the basis of two-stream architectures. It can be proved by the experiments in \autoref{sec:4} that the improved deep learning based classification methods are more competitive than ordinary methods. The relationship between MCNN and other commonly used architectures can be seen from \autoref{fig_venn}.
\par The proposed MCNN draws on a lot of expert experience from existing CNNs so there are many similarities with common architectures. Like some recent architectures, MCNN is chain-structured and on basis of mainstream backbones. We are greatly inspired by the two-stream CNNs in process of modeling, which can be seen from the comparison between \autoref{fig1.1} and \autoref{fig4}. For more powerful performance, side outputs are utilized to implement advanced information fusion by deeply-supervised skip connections and multi-task architectures. Therefore, the architectures adopted in this work intersect with three other ones, as shown in the Venn diagram.
\par It is wroth to note that the proposed method is quite different from the common ones. The main reason for the differences lies in the changes of the input. Since the starting point of our modeling is to better adapt to PolSAR data, some targeted improvements have been implemented based on some existing tricks to achieve an easy-optimized and high-performance architecture. Firstly, two branches exist in the proposed architecture which are used to process two kinds of data respectively. This is the targeted adjustment we made to fit the used input form. Then, the proposed architecture is equipped with the more carefully designed information fusion mechanisms compared to commonly used two-stream CNNs. A densely connected fusion module is used to fuse the information obtained by the two branches while maintaining the acquired features. Further, there are three extra classifiers in MCNN. The significance of their existence can be divided into two aspects. On one hand, they are beneficial to the network optimization. It is difficult to optimize a deep network architecture, especially under the context of PolSAR image classification, which is a task with poor data diversity. Regularization methods are common techniques to alleviate this problem, such as ReLU activation, batch normalization and skip connections \citep{dpen}. Additional classifiers bring the additional task of enhancing the presentation ability of the side outputs, which can also be seen as a variant of regularization. On the other hand, better predictions can be achieved through the proposed advanced fusion mechanism, which are on basis of the classification results obtained by the extra classifiers. Based on the theoretical analyses, components for improvements are designed and introduced. Ablation experiments, as shown in \autoref{figcompare}, show the validity of each decomposition. As a result, the proposed architecture becomes different from the common ones.
\section{Conclusion}
\label{sec:5}
In this work, a PolSAR tailored classification framework, named multi-task convolutional neural network (MCNN), is built to release the potential of deep learning techniques in PolSAR image classification. The need of adjusting CNN architectures according to the characteristics of input data is fully considered in the process of modeling. The construction of the proposed classification framework can be mainly divided into two parts. Firstly, input form of complex-valued PolSAR data is changed to its amplitude and phase instead of the commonly used real and imaginary parts. Secondly, a novel MCNN architecture is built to better extract features from the amplitude and phase, which integrates the idea of two-stream CNNs and multi-task models. Different paths are defined in the proposed MCNN to extract distinctive information for different types of information of the input. Multi-level information interaction mechanism is applied to the proposed architecture to achieve more comprehensive classification results. Further, in order to better model the potential correlations from the phase of PolSAR images, depthwise separable convolution is introduced into MCNN and a depthwise separable convolution based architecture, named DMCNN, is constructed to better utilize the phase information and improve the classification results.
\par Experiments on three widely used PolSAR benchmark datasets show that the proposed framework has certain advantages over ordinary input form and optical image classification used CNN architectures.
\par For the future works, better input forms of PolSAR data, well-designed CNN architectures and their application to low-shot learning, and especially the application of neural architecture search methods in PolSAR area are all the issues we are considering.\\
\noindent {\bf{Acknowledgements}}\\
This work was supported in part by the National Natural Science Foundation of China (61401124, 61871158), in part by Scientific Research Foundation for the Returned Overseas Scholars of Heilongjiang Province (LC2018029), in part by Aeronautical Science Foundation of China (20182077008).

\bibliographystyle{model2-names}

\bibliography{References}





\end{document}